\documentclass[11pt]{article}
\usepackage[utf8]{inputenc}
\usepackage[T1]{fontenc}
\usepackage[a4paper,margin=1in]{geometry}
\usepackage{textcomp}
\usepackage{booktabs}
\usepackage{multirow}
\usepackage{graphicx}
\usepackage{amsmath}
\usepackage{longtable,tabularx}
\usepackage{amssymb}
\usepackage{algorithm}
\usepackage{algpseudocode}
\usepackage[numbers,sort&compress]{natbib}
\usepackage[hidelinks]{hyperref}
\usepackage{doi}
\usepackage{microtype}
\usepackage[hidelinks]{hyperref}
\usepackage{url}

\title{Hybrid Machine Learning for Imbalanced Landing-Phase Safety Event Prediction from Flight Data}

\author{%
Taha Asghar$^{1}$, Bilal Mufti$^{1}$,\\
Syed Qasim Zaheer$^{1}$, and Luqman Ahmed$^{1}$\\[0.5em]
\small $^{1}$College of Aeronautical Engineering,\\
\small National University of Sciences and Technology (NUST),\\
\small Risalpur, Pakistan\\[0.5em]
\small Corresponding author: Bilal Mufti,
\texttt{bmufti@cae.nust.edu.pk}}
\date{}

\begin{document}

\maketitle

\begin{abstract}
Landing-phase safety-event prediction requires the detection of rare
abnormal approaches within strongly imbalanced multivariate flight
data. Capturing coupled temporal behavior while distinguishing
event-specific signatures across airport operating contexts remains
a central modeling challenge. This study proposes an interpretable
hybrid feature-fusion framework for multiclass landing-phase
safety-event prediction. A long short-term memory (LSTM) autoencoder
trained on nominal flight windows characterizes deviations from
routine approach behavior. Its continuous reconstruction-error
descriptor is fused with parameter-specific statistical flight-history
features and supplied to a class-weighted XGBoost classifier to
distinguish Normal, Speed High, Path High, and Flaps Late windows.
The framework is evaluated using 53,750 NASA DASHlink approach
windows from three arrival airports. Pooled and airport-specific
learning strategies are compared alongside component-level
benchmarks, repeated pipeline retraining, and feature-attribution
analysis. The results demonstrate strong class-balanced detection,
with mean macro recall above 0.90 across all airports under repeated
partitioning and retraining. Component comparisons establish the
importance of class weighting and show that continuous feature
fusion outperforms reconstruction-error-only and hard-gate
alternatives. Airport-level comparisons reveal benefits from local
specialization at the two airports with larger datasets and from
pooled training at the airport with the smallest dataset. Feature
attributions associate predictions with operationally relevant
vertical-path, configuration, speed, and approach-context variables.
The framework provides an interpretable approach to rare-event
detection and practical guidance for adapting flight-data monitoring
models to airport-level data availability.
\end{abstract}

% \section*{Nomenclature}

% \noindent(Nomenclature entries should have the units identified)

% {\renewcommand\arraystretch{1.0}
% \noindent\begin{longtable*}{@{}l @{\quad=\quad} l@{}}
% $A$  & amplitude of oscillation \\
% $a$ &    cylinder diameter \\
% $C_p$& pressure coefficient \\
% $Cx$ & force coefficient in the \textit{x} direction \\
% $Cy$ & force coefficient in the \textit{y} direction \\
% c   & chord \\
% d$t$ & time step \\
% $Fx$ & $X$ component of the resultant pressure force acting on the vehicle \\
% $Fy$ & $Y$ component of the resultant pressure force acting on the vehicle \\
% $f, g$   & generic functions \\
% $h$  & height \\
% $i$  & time index during navigation \\
% $j$  & waypoint index \\
% $K$  & trailing-edge (TE) nondimensional angular deflection rate\\
% $\Theta$ & boundary-layer momentum thickness\\
% $\rho$ & density\\
% \multicolumn{2}{@{}l}{Subscripts}\\
% cg & center of gravity\\
% $G$ & generator body\\
% iso	& waypoint index
% \end{longtable*}}

% \addtocounter{table}{-1}

\section{Introduction}

Commercial aviation has achieved a high level of safety over the past several decades, despite long-term growth in air transport activity~\cite{boeing2024statsum,faa2017forecast}. This improvement has been supported by advances in aircraft technology, regulatory oversight, operator safety-management practices, crew training, and the systematic use of operational data. Boeing's statistical summary of commercial jet airplane accidents reports substantial reductions in total and fatal accident rates between the two most recent decades covered by the report, alongside an increase in departures~\cite{boeing2024statsum}. Continued vigilance remains necessary as traffic volume and operational complexity increase. Data-driven analysis can support this effort by identifying operational patterns relevant to risk assessment~\cite{hegde2020applications}.

Approach and landing remain particularly safety-sensitive parts of flight. During these phases, airspeed control, vertical-path tracking, heading alignment, and aircraft configuration must be coordinated within limited recovery margins near touchdown. Boeing's commercial jet accident statistics indicate that landing accounts for only a small fraction of total flight time, but contributes a disproportionately large share of fatal accidents~\cite{boeing2024statsum}. Safety studies and reports associate unstable approaches and deviations in flight state with adverse outcomes, including hard landings, runway excursions, and tail strikes~\cite{li2023overview,iata2023accidentprevention}. These deviations can develop over short time intervals and may involve coupled changes across multiple flight parameters. Routine recording of flight-state and control parameters~\cite{faa2011dfdr} provides multivariate histories for examining such behavior.

Flight Operational Quality Assurance (FOQA), closely related to flight-data monitoring and flight-data analysis, uses routinely recorded aircraft data to support safety management. In a FOQA program, flight-recorded parameters are reviewed to identify exceedances, monitor operational trends, and support non-punitive safety improvement activities~\cite{faa2004foqa,icao_fdap,caa_cap739}. For approach-and-landing analysis, these programs allow deviations in airspeed, vertical path, aircraft configuration, and control response to be examined together. Traditional workflows commonly use predefined exceedance thresholds and post-flight review~\cite{faa2004foqa,haas2008using}. These workflows can be complemented by data-driven models that learn recurring patterns associated with safety-relevant operational states from multivariate flight histories.

Learning these patterns presents three coupled challenges. First, the data are strongly imbalanced because nominal operations occur far more frequently than safety-relevant events; this imbalance can cause models to achieve high overall accuracy while failing to detect rare but operationally important classes~\cite{gao2025imbalanced,bleu2022multiclass}. Second, the relevant information is distributed across time and across multiple flight parameters. During approach, speed management, glide-path tracking, configuration timing, wind effects, and control-surface response are coupled, so event signatures may appear as multivariate temporal patterns rather than isolated exceedances in a single variable~\cite{bleu2022multiclass}. Third, differences in weather exposure, runway characteristics, and local operating procedures across airports motivate evaluating whether pooled and airport-specific models behave differently~\cite{midtfjord2022runway}. These factors motivate a prediction framework that combines temporal pattern learning, compact statistical representation, imbalance-aware classification, and explicit evaluation of airport-specific behavior.

This study addresses these challenges by developing a hybrid feature-fusion framework for landing-phase flight safety event prediction. A long short-term memory (LSTM) autoencoder is first used to learn nominal temporal patterns and generate a reconstruction-error descriptor for each flight window. This descriptor is combined with statistical features that summarize the level, variability, shape, and trends of the recorded flight-parameter histories. The resulting fixed-length representation is provided to a class-weighted XGBoost classifier for multiclass event prediction, reducing the tendency of the model to favor the dominant nominal class. The study evaluates the framework under both pooled and airport-specific training strategies, showing strong event-detection performance while identifying differences in the relative performance of the two strategies across airports. The study contributes a framework for imbalanced landing-phase safety-event prediction and provides empirical evidence on how temporal-deviation information, class weighting, and airport-specific learning affect the detection of underrepresented safety events.

\section{Background and Related Work}
\label{sec:background_related_work}

Data-driven modeling has become an important complement to physics-based analysis in aerospace applications, particularly when repeated first-principles analysis of high-dimensional problems is expensive or impractical~\cite{li2022machine,brunton2020machine,mufti2024shock,mufti2025nonlinear}. In aviation safety, flight-data monitoring and related safety-data programs have established the use of routinely recorded operational data for proactive risk identification, safety assurance, and trend monitoring~\cite{faa2004foqa,oehling2019using,rey2021data}. Reviews and application studies across reliability engineering, transportation safety, and aircraft hazard prediction highlight the use of machine learning to extract safety-relevant patterns from complex operational data~\cite{xu2020machine,silva2020machine,zhou2020deep}. Within this context, Puranik et al.~\cite{puranik2020online} developed an offline--online supervised-learning framework for predicting safety-critical landing metrics, including touchdown true airspeed and ground speed, from approach-phase data extracted between operational decision gates. Their study demonstrated the value of supervised learning for approach-and-landing monitoring, but the prediction target was a continuous landing-performance metric rather than a discrete safety-event class. Wang and Mou~\cite{wang2025flight} addressed approach-and-landing risk prediction from a flight-state perspective by selecting indicators from longitudinal and lateral motion analysis and training a LightGBM-based risk model. Bleu Laine et al.~\cite{bleu2022multiclass} examined multiclass adverse-event prediction and precursor identification using multivariate flight histories. These studies establish the relevance of supervised learning for approach-and-landing safety analysis, while also showing that the literature spans distinct problem formulations: landing-metric regression, flight-state risk prediction, and multiclass adverse-event classification.

A central difficulty in flight-data-based safety prediction is that approach-and-landing events are embedded in multivariate time histories rather than isolated parameter values. Flight-state deviations during approach and landing are associated with coupled changes in speed, altitude, attitude, path tracking, configuration, and control response~\cite{wang2025flight,puranik2020online,li2024imtcn}. Sequence-learning models provide one way to represent such coupled temporal behavior. Long short-term memory networks, for example, were developed to capture long-range dependencies in sequential data~\cite{hochreiter1997lstm,gers2000learning,greff2017lstm}. Autoencoders can learn a compact representation of nominal behavior, with sequence-based variants capturing its temporal structure; reconstruction error then provides a measure of deviation from the learned pattern~\cite{sakurada2014anomaly,malhotra2016lstm,pang2021deep,chalapathy2019deep}. In aviation safety, Bleu Laine et al.~\cite{bleu2022multiclass} used a multihead CNN--RNN architecture in which individual sensor histories were first processed separately and then combined to learn temporal interactions for adverse-event prediction and precursor identification. Li et al.~\cite{li2024imtcn} formulated flight exceedance prediction as a multiscale time-series classification problem and used temporal-convolutional modules to capture local temporal features and long effective histories from quick access recorder data. For UAV anomaly detection, Li et al.~\cite{li2026tsae} developed a time-sense autoencoder that models multiscale periodic patterns in flight histories. For cruise-phase monitoring, Turgut~\cite{turgut2026phase} combined perceptual hashing and Shannon entropy to characterize profile deviations and signal instabilities in multivariate quick access recorder data. A complementary strategy is to summarize each flight window into a fixed-length representation and use a supervised classifier to learn event boundaries in this representation. XGBoost, a regularized gradient-boosted decision-tree method, is well suited to these fixed-length feature spaces because it can capture nonlinear feature interactions and has shown strong performance in tabular prediction problems~\cite{friedman2001greedy,chen2016xgboost}. Applications of boosted-tree models to runway-condition assessment and approach-and-landing risk prediction also demonstrate their relevance to modeling engineered operational features~\cite{midtfjord2022runway,wang2025flight}. Mao et al.~\cite{mao2025flighttechnique} combined ensemble learning with resampling, adaptive dimensionality reduction, and hyperparameter optimization for flight-technique assessment, using SHAP to examine the contributions of individual flight parameters.

A second major challenge is the imbalance between nominal operations and safety-relevant events. In their DASHlink-based adverse-event prediction study, Bleu Laine et al.~\cite{bleu2022multiclass} selected severity-level-3 events and included a subset of nominal flights matched to the combined number of adverse-event cases. The severity restriction limits overlap between nominal and abnormal operations, while nominal-flight subsampling reduces class imbalance. The resulting evaluation setting differs from the strongly imbalanced distributions considered here. In aircraft sensor-anomaly detection, Gao et al.~\cite{gao2025imbalanced} showed that class imbalance can impair detection and investigated resampling and loss modification to improve minority-class performance. For rare flight-diversion prediction, Aly et al.~\cite{aly2026diversion} developed a generative augmentation framework for imbalanced tabular aviation records, jointly considering synthetic-data quality and predictive utility. More generally, imbalanced-learning studies have shown that sampling strategies, cost-sensitive learning, and class-weighted losses can improve detection when minority classes carry greater practical importance than their sample frequency suggests~\cite{he2009learning,chawla2002smote,fernandez2018learning}. For landing-phase safety-event prediction, this means that overall accuracy is not sufficient: a model can perform well on the dominant Normal class while still missing rare but operationally important events. Class-sensitive metrics such as macro recall and macro F1 are therefore more informative for evaluating multiclass safety-event prediction under imbalance.

Local operating context introduces an additional modeling and interpretation challenge. Approach-and-landing behavior depends on aircraft state as well as runway configuration, approach procedures, traffic constraints, weather exposure, surface condition, and other airport-specific factors. Puranik et al.~\cite{puranik2020online} trained a global landing-performance model using data from multiple airframes and more than seventy airports, demonstrating the applicability of pooled learning across a broad operational dataset. Their formulation focused on continuous touchdown-speed prediction and did not compare pooled and airport-specific models for multiclass safety-event prediction. Midtfjord et al.~\cite{midtfjord2022runway} noted that airports differ in weather conditions, maintenance procedures, runway lengths, and traffic, and therefore restricted their analysis to a single airport to avoid confounding effects from airport-to-airport variation. These studies suggest that pooled and airport-specific models may offer different advantages: pooled models benefit from larger and more diverse training data, whereas airport-specific models may better represent local operating signatures. Comparing these models also raises an interpretability question: whether the variables driving predictions remain operationally meaningful across airports and event classes. Interpretable flight-safety models such as IMTCN identify key flight parameters and time instants associated with exceedance predictions~\cite{li2024imtcn}. For tree-based models, SHAP and Tree SHAP provide feature-attribution values for ensemble decision-tree predictions~\cite{lundberg2017unified,lundberg2020local,midtfjord2022runway}. These explanations let us examine the variables associated with model predictions in relation to their operational meaning, while recognizing that the attributions describe the trained model's decision behavior rather than the underlying physical mechanism.

These challenges occur together in multiclass landing-phase safety-event prediction: event signatures are distributed across multivariate time histories, rare event classes can be masked by dominant nominal operations, and local operating context can change the observed patterns associated with the same event label. This motivates examining how temporal-deviation information complements statistical flight-history descriptors under class imbalance, and how pooled and airport-specific learning affect minority-event detection. The present study examines these questions using an LSTM-autoencoder reconstruction-error descriptor, engineered statistical features, and a class-weighted XGBoost classifier. The evaluation compares pooled and airport-specific performance, examines class-wise errors and component-level contributions, and uses SHAP to interpret airport- and event-specific feature drivers.

\section{Methodology}
\label{sec:methodology}

This section presents the formulation and implementation of the proposed methodology. The complete hybrid feature-fusion framework is first described, followed by the LSTM-autoencoder reconstruction-error descriptor, statistical flight-history feature extraction, and class-weighted XGBoost event classification.

\subsection{Proposed Hybrid Feature-Fusion Framework}
\label{subsec:hybrid_framework}

\begin{figure*}[t]
    \centering
    \includegraphics[width=\textwidth]
    {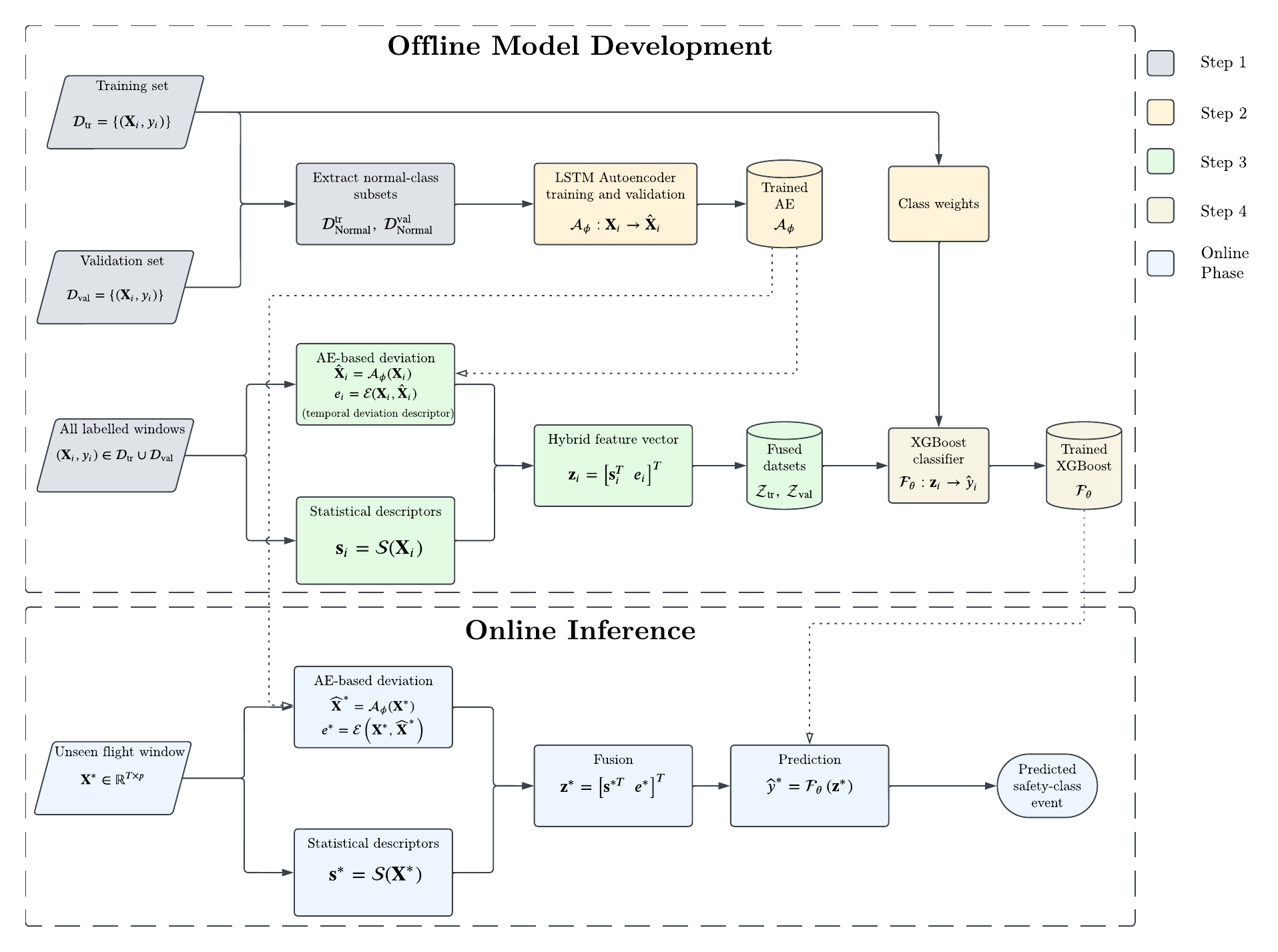}
    \caption{Proposed hybrid feature-fusion framework for landing-phase safety-event prediction. Statistical flight-history descriptors and an LSTM-autoencoder reconstruction-error descriptor are extracted from the same multivariate flight window, concatenated, and supplied to a class-weighted XGBoost classifier for multiclass prediction.}
    \label{fig:hybrid_framework}
\end{figure*}

The proposed framework represents each preprocessed flight window using two complementary feature-extraction branches, as illustrated in Fig.~\ref{fig:hybrid_framework}. Let the $i$th multivariate flight window be denoted by
\begin{equation}
    \mathbf{X}_i =
    \begin{bmatrix}
        \mathbf{x}_i^{(1)} &
        \mathbf{x}_i^{(2)} &
        \cdots &
        \mathbf{x}_i^{(p)}
    \end{bmatrix}
    \in \mathbb{R}^{T \times p},
    \label{eq:flight_window}
\end{equation}
where $T$ denotes the number of temporal samples, $p$ is the number of recorded flight parameters, and $\mathbf{x}_i^{(j)}$ represents the time history of the $j$th parameter. The same flight window is supplied to a statistical feature-extraction branch and an LSTM-autoencoder branch to represent parameter-specific flight-history characteristics alongside deviations from learned nominal temporal behavior.

In the statistical branch, an extraction operator $\mathcal{S}(\cdot)$ transforms the multivariate flight window into a fixed-length descriptor vector,
\begin{equation}
    \mathbf{s}_i = \mathcal{S}(\mathbf{X}_i)
    \in \mathbb{R}^{d_s},
    \label{eq:statistical_vector}
\end{equation}
where $d_s$ denotes the dimension of the resulting statistical representation. These descriptors summarize characteristics of the individual parameter histories while preserving the association between each descriptor and its parent flight parameter.

In parallel, the LSTM autoencoder reconstructs the input sequence as
\begin{equation}
    \widehat{\mathbf{X}}_i
    =
    \mathcal{A}_{\boldsymbol{\phi}}(\mathbf{X}_i),
    \label{eq:ae_reconstruction}
\end{equation}
where $\mathcal{A}_{\boldsymbol{\phi}}(\cdot)$ denotes the trained autoencoder with parameters $\boldsymbol{\phi}$. A reconstruction-error operator $\mathcal{E}(\cdot)$ is then used to obtain a scalar descriptor,
\begin{equation}
    e_i =
    \mathcal{E}
    \left(
        \mathbf{X}_i,
        \widehat{\mathbf{X}}_i
    \right),
    \label{eq:reconstruction_error_generic}
\end{equation}
which quantifies the overall deviation of the observed multivariate sequence from the nominal temporal behavior learned by the autoencoder. The reconstruction error is retained as a continuous feature rather than thresholded to assign a window to the Normal class or a safety-event class.

The two representations are concatenated to form the hybrid feature vector
\begin{equation}
    \mathbf{z}_i =
    \begin{bmatrix}
        \mathbf{s}_i^{\mathsf{T}} & e_i
    \end{bmatrix}^{\mathsf{T}}
    \in \mathbb{R}^{d_s+1}.
    \label{eq:fused_feature_vector}
\end{equation}
The fused representation is supplied to a class-weighted XGBoost classifier,
\begin{equation}
    \widehat{y}_i =
    \mathcal{F}_{\boldsymbol{\theta}}(\mathbf{z}_i),
    \qquad
    \widehat{y}_i \in \{1,\ldots,C\},
    \label{eq:multiclass_prediction}
\end{equation}
where $\mathcal{F}_{\boldsymbol{\theta}}(\cdot)$ denotes the trained classifier, $\boldsymbol{\theta}$ represents its learned parameters, and $C$ is the number of prediction classes. Class weighting is incorporated during training to reduce the tendency to favor classes with greater sample support.

The model-development and training procedure is summarized in Algorithm~\ref{alg:hybrid_framework}. The preprocessing transformations are first determined from the training data, after which the autoencoder is trained using the nominal flights training subset. Statistical descriptors and reconstruction errors are then generated for the labelled training and validation windows and concatenated to form the classifier inputs. The training features are used to fit the class-weighted classifier, while the validation features support model selection. For an unseen preprocessed flight window, the statistical feature extractor and trained autoencoder generate the corresponding descriptors, which are concatenated and supplied to the trained XGBoost classifier for multiclass prediction.

\begin{algorithm}[t]
\caption{Development and training of the proposed hybrid
feature-fusion framework}
\label{alg:hybrid_framework}
\begin{algorithmic}[1]

\Require Labelled, preprocessed training and validation sets
$\mathcal{D}_{\mathrm{tr}}$ and $\mathcal{D}_{\mathrm{val}}$,
each containing the Normal class and the safety-event classes

\Ensure Trained hybrid prediction framework
$\mathcal{H}
=
\{\mathcal{S},
\mathcal{A}_{\boldsymbol{\phi}},
\mathcal{F}_{\boldsymbol{\theta}}\}$

\Statex
\Statex \textbf{Stage 1: Learn nominal temporal behavior}

\State Extract the Normal-class subsets
$\mathcal{D}_{\mathrm{Normal}}^{\mathrm{tr}}
\subset \mathcal{D}_{\mathrm{tr}}$
and
$\mathcal{D}_{\mathrm{Normal}}^{\mathrm{val}}
\subset \mathcal{D}_{\mathrm{val}}$

\State Train the LSTM autoencoder
$\mathcal{A}_{\boldsymbol{\phi}}$
using
$\mathcal{D}_{\mathrm{Normal}}^{\mathrm{tr}}$

\State Select the autoencoder configuration using reconstruction
performance on
$\mathcal{D}_{\mathrm{Normal}}^{\mathrm{val}}$

\State Freeze the trained autoencoder parameters
$\boldsymbol{\phi}$

\Statex
\Statex \textbf{Stage 2: Construct the fused feature representation}

\ForAll{$q \in \{\mathrm{tr},\mathrm{val}\}$}

    \State Initialize the fused dataset
    $\mathcal{Z}_{q} \gets \varnothing$

    \ForAll{$(\mathbf{X}_i,y_i)\in\mathcal{D}_{q}$}

        \State Extract the statistical flight-history descriptors:
        $\mathbf{s}_i \gets \mathcal{S}(\mathbf{X}_i)$

        \State Reconstruct the multivariate flight window:
        $\widehat{\mathbf{X}}_i
        \gets
        \mathcal{A}_{\boldsymbol{\phi}}(\mathbf{X}_i)$

        \State Calculate the reconstruction-error descriptor:
        $e_i
        \gets
        \mathcal{E}
        (\mathbf{X}_i,\widehat{\mathbf{X}}_i)$

        \State Concatenate the two representations:
        $\mathbf{z}_i
        \gets
        [\mathbf{s}_i^{\mathsf{T}},e_i]^{\mathsf{T}}$

        \State Add the fused vector and its original class label:
        $\mathcal{Z}_{q}
        \gets
        \mathcal{Z}_{q}\cup\{(\mathbf{z}_i,y_i)\}$

    \EndFor

\EndFor

\Statex
\Statex \textbf{Stage 3: Train the imbalance-aware classifier}

\State Determine class weights from the class frequencies in
$\mathcal{D}_{\mathrm{tr}}$

\State Train candidate XGBoost classifiers on
$\mathcal{Z}_{\mathrm{tr}}$
using the class weights

\State Select the final classifier
$\mathcal{F}_{\boldsymbol{\theta}}$
using performance on
$\mathcal{Z}_{\mathrm{val}}$

\State Combine the statistical feature extractor,
trained autoencoder, and selected classifier into the complete
hybrid framework

\Return
$\mathcal{H}
=
\{\mathcal{S},
\mathcal{A}_{\boldsymbol{\phi}},
\mathcal{F}_{\boldsymbol{\theta}}\}$

\end{algorithmic}
\end{algorithm}

\subsection{LSTM-Autoencoder Reconstruction-Error Descriptor}
\label{subsec:lstm_autoencoder}

The temporal branch of the proposed framework uses an LSTM autoencoder to characterize deviations from the temporal behavior observed in the Normal class. The autoencoder is trained to reconstruct its input without using the safety-event labels as prediction targets. Restricting its training data to Normal-class flight windows encourages the model to learn a compact representation of routine multivariate approach behavior.

Let
\begin{equation}
    \mathbf{X}_i =
    \left[
        \mathbf{x}_{i,1},
        \mathbf{x}_{i,2},
        \ldots,
        \mathbf{x}_{i,T}
    \right]^{\mathsf{T}}
    \in \mathbb{R}^{T\times p}
\end{equation}
denote the $i$th multivariate flight window, where
$\mathbf{x}_{i,t}\in\mathbb{R}^{p}$ contains the recorded flight
parameters at time index $t$. The encoder processes the sequence and
maps it to a latent representation,
\begin{equation}
    \mathbf{h}_i
    =
    \mathcal{E}_{\boldsymbol{\phi}_{e}}(\mathbf{X}_i),
    \label{eq:lstm_encoder}
\end{equation}
where $\mathcal{E}_{\boldsymbol{\phi}_{e}}(\cdot)$ denotes the
LSTM-based encoder. The decoder then expands this representation to
reconstruct the complete input sequence,
\begin{equation}
    \widehat{\mathbf{X}}_i
    =
    \mathcal{D}_{\boldsymbol{\phi}_{d}}(\mathbf{h}_i),
    \label{eq:lstm_decoder}
\end{equation}
where $\mathcal{D}_{\boldsymbol{\phi}_{d}}(\cdot)$ denotes the
LSTM-based decoder. The complete autoencoder is therefore written as
\begin{equation}
    \widehat{\mathbf{X}}_i
    =
    \mathcal{A}_{\boldsymbol{\phi}}(\mathbf{X}_i)
    =
    \mathcal{D}_{\boldsymbol{\phi}_{d}}
    \left(
        \mathcal{E}_{\boldsymbol{\phi}_{e}}(\mathbf{X}_i)
    \right),
    \qquad
    \boldsymbol{\phi}
    =
    \{\boldsymbol{\phi}_{e},\boldsymbol{\phi}_{d}\}.
    \label{eq:lstm_autoencoder}
\end{equation}

The autoencoder parameters are learned by minimizing reconstruction
loss over the Normal-class training subset
$\mathcal{D}_{\mathrm{Normal}}^{\mathrm{tr}}$:
\begin{equation}
    \boldsymbol{\phi}^{*}
    =
    \arg\min_{\boldsymbol{\phi}}
    \frac{1}
    {\left|
        \mathcal{D}_{\mathrm{Normal}}^{\mathrm{tr}}
    \right|}
    \sum_{\mathbf{X}_i
    \in
    \mathcal{D}_{\mathrm{Normal}}^{\mathrm{tr}}}
    \mathcal{L}_{\mathrm{rec}}
    \left(
        \mathbf{X}_i,
        \mathcal{A}_{\boldsymbol{\phi}}(\mathbf{X}_i)
    \right).
    \label{eq:ae_training_objective}
\end{equation}
The reconstruction loss is the mean squared error across all
temporal samples and flight parameters:
\begin{equation}
    \mathcal{L}_{\mathrm{rec}}
    \left(
        \mathbf{X}_i,
        \widehat{\mathbf{X}}_i
    \right)
    =
    \frac{1}{Tp}
    \sum_{t=1}^{T}
    \sum_{j=1}^{p}
    \left(
        X_{i,tj}
        -
        \widehat{X}_{i,tj}
    \right)^{2}.
    \label{eq:ae_reconstruction_loss}
\end{equation}

After training, the autoencoder parameters are fixed and the model is
applied to every labelled flight window, including both Normal and
safety-event classes. For each window, the scalar reconstruction-error
descriptor is computed as
\begin{equation}
    e_i
    =
    \frac{1}{Tp}
    \left\|
        \mathbf{X}_i
        -
        \widehat{\mathbf{X}}_i
    \right\|_{F}^{2},
    \label{eq:ae_error_descriptor}
\end{equation}
where $\|\cdot\|_{F}$ denotes the Frobenius norm. A larger value of
$e_i$ indicates a greater reconstruction discrepancy relative to the
nominal temporal behavior learned from Normal-class windows.

In the proposed framework, $e_i$ is concatenated with the statistical
flight-history features as a continuous temporal-deviation descriptor,
without applying a hard anomaly threshold. The downstream classifier
interprets reconstruction error jointly with parameter-specific
statistical information when distinguishing among the safety-event
classes.

\subsection{Statistical Flight-History Feature Extraction}
\label{subsec:statistical_features}

The second branch of the proposed framework converts each multivariate
flight window into a fixed-length statistical representation. This
representation complements the autoencoder-based deviation descriptor
by retaining parameter-specific information about the magnitude,
dispersion, distributional shape, and temporal trends of the recorded
flight histories.

For the $i$th flight window, let
$\mathbf{x}_{i}^{(j)}
=
[x_{i,1}^{(j)},\ldots,x_{i,T}^{(j)}]^{\mathsf T}$
denote the time history of the $j$th flight parameter. A set of
$K$ statistical descriptors is calculated independently for each
parameter:
\begin{equation}
    \mathbf{g}_{i}^{(j)}
    =
    \begin{bmatrix}
        g_{1}\!\left(\mathbf{x}_{i}^{(j)}\right) &
        \cdots &
        g_{K}\!\left(\mathbf{x}_{i}^{(j)}\right)
    \end{bmatrix}^{\mathsf T},
    \label{eq:sensor_statistical_vector}
\end{equation}
where $g_k(\cdot)$ denotes the $k$th descriptor. The descriptor set used
in the present framework is summarized in
Table~\ref{tab:statistical_descriptors}.

\begin{table}[t]
\centering
\caption{Statistical descriptors extracted from each flight-parameter
history.}
\label{tab:statistical_descriptors}
\begin{tabular}{ll}
\toprule
\textbf{Descriptor} & \textbf{Purpose} \\
\midrule
Mean & Average parameter level \\
Standard deviation & Overall variability \\
Maximum & Upper extreme \\
Minimum & Lower extreme \\
Range & Peak-to-peak variation \\
Linear slope & Direction and rate of temporal change \\
Correlation coefficient & Strength of the linear temporal trend \\
Median & Robust central tendency \\
Interquartile range & Robust measure of dispersion \\
Skewness & Distributional asymmetry \\
Kurtosis & Distributional shape and tail weight \\
\bottomrule
\end{tabular}
\end{table}

The temporal-trend descriptors are obtained from a least-squares linear
fit of each parameter history against the sample index,
\begin{equation}
    x_{i,t}^{(j)}
    =
    a_{i}^{(j)}
    +
    b_{i}^{(j)} t
    +
    \epsilon_{i,t}^{(j)},
    \label{eq:linear_trend_fit}
\end{equation}
where $b_{i}^{(j)}$ is retained as the slope descriptor. The correlation
coefficient between the parameter history and the sample index is also
included to indicate the strength of the linear trend. Together, these
descriptors distinguish histories with similar slopes but different
degrees of linear temporal association.

The descriptor vectors from all $p$ flight parameters are concatenated
to form the statistical feature representation,
\begin{equation}
    \mathbf{s}_i
    =
    \begin{bmatrix}
        \mathbf{g}_{i}^{(1)\mathsf T} &
        \mathbf{g}_{i}^{(2)\mathsf T} &
        \cdots &
        \mathbf{g}_{i}^{(p)\mathsf T}
    \end{bmatrix}^{\mathsf T}
    \in \mathbb{R}^{pK}.
    \label{eq:statistical_feature_vector}
\end{equation}
For the implementation considered in this study, $K=11$ descriptors are
calculated for each of the $p=20$ flight parameters, producing
\begin{equation}
    d_s=pK=20\times11=220
    \label{eq:statistical_feature_dimension}
\end{equation}
statistical features for each flight window.

This fixed-length representation provides the tree-based classifier
with parameter-specific summaries of flight-state levels, variability,
extrema, distributional shape, and temporal trends, without treating
every recorded time sample as a separate input feature. It reduces
the dimensionality of the classifier input while preserving the
association between each descriptor and its parent flight parameter,
which also supports subsequent model interpretation. These summaries
do not retain the detailed temporal ordering of the original histories;
the autoencoder branch complements them by measuring reconstruction
discrepancy across the multivariate sequence. The statistical vector
$\mathbf{s}_i$ is concatenated with the scalar autoencoder
reconstruction-error descriptor to construct the final hybrid feature
vector.

\subsection{Class-Weighted XGBoost Event Classification}
\label{subsec:weighted_xgboost}

The final stage of the proposed framework performs supervised
multiclass classification using the fused statistical and
temporal-deviation representation. For the $i$th flight window, the
statistical feature vector $\mathbf{s}_i$ is concatenated with the
scalar autoencoder reconstruction-error descriptor $e_i$ to form
\begin{equation}
    \mathbf{z}_i
    =
    \begin{bmatrix}
        \mathbf{s}_i^{\mathsf T} & e_i
    \end{bmatrix}^{\mathsf T}
    \in \mathbb{R}^{d_s+1}.
    \label{eq:hybrid_feature_vector}
\end{equation}
For the present implementation, $d_s=220$, giving a
$221$-dimensional hybrid feature vector. This representation combines
parameter-specific statistical information with a compact measure of
deviation from the nominal temporal behavior learned by the LSTM
autoencoder.

XGBoost represents the multiclass decision function as an additive
ensemble of regression trees. At boosting iteration $m$, the
class-score vector is updated according to
\begin{equation}
    \mathbf{f}^{(m)}(\mathbf{z}_i)
    =
    \mathbf{f}^{(m-1)}(\mathbf{z}_i)
    +
    \eta\,\mathbf{h}_{m}(\mathbf{z}_i),
    \label{eq:xgboost_additive_model}
\end{equation}
where $\mathbf{h}_{m}(\cdot)$ denotes the vector of contributions
from the class-specific trees added at iteration $m$, and $\eta$ is
the learning rate. The resulting class scores are converted to class
probabilities using the softmax function,
\begin{equation}
    P(y_i=c\mid\mathbf{z}_i)
    =
    \frac{
        \exp\!\left(f_c(\mathbf{z}_i)\right)
    }{
        \displaystyle
        \sum_{r=1}^{C}
        \exp\!\left(f_r(\mathbf{z}_i)\right)
    },
    \qquad c=1,\ldots,C,
    \label{eq:xgboost_softmax}
\end{equation}
and the predicted class is obtained as
\begin{equation}
    \widehat{y}_i
    =
    \arg\max_{c\in\{1,\ldots,C\}}
    P(y_i=c\mid\mathbf{z}_i).
    \label{eq:xgboost_prediction}
\end{equation}

Because the training data are strongly imbalanced, assigning equal
importance to every training sample can cause the objective to be
dominated by the majority class. A class weight is therefore calculated
from the training-set class frequencies as
\begin{equation}
    w_c
    =
    \frac{N_{\mathrm{tr}}}
    {C\,n_c},
    \label{eq:class_weight}
\end{equation}
where $N_{\mathrm{tr}}$ is the total number of training samples,
$C$ is the number of classes, and $n_c$ is the number of training
samples belonging to class $c$. Each sample inherits the weight of its
class:
\begin{equation}
    \omega_i=w_{y_i}.
    \label{eq:sample_weight}
\end{equation}
The weighted training objective is expressed as
\begin{equation}
    \mathcal{J}
    =
    \sum_{i=1}^{N_{\mathrm{tr}}}
    \omega_i\,
    \ell
    \left(
        y_i,
        \mathbf{f}(\mathbf{z}_i)
    \right)
    +
    \sum_{m=1}^{M}
    \Omega(\mathbf{h}_m),
    \label{eq:weighted_xgboost_objective}
\end{equation}
where $\ell(\cdot)$ is the multiclass classification loss,
$\Omega(\mathbf{h}_m)$ denotes the total complexity penalty for the
trees added at iteration $m$, and $M$ is the number of boosting
iterations. The weighting term increases the contribution of
underrepresented classes without resampling the training data.

The classifier is trained using the fused representations and original
class labels from the training partition, while the validation
partition is used for model selection. During inference, the
statistical feature extractor and trained autoencoder generate the
hybrid representation for an unseen preprocessed flight window.
The fixed XGBoost model then uses this representation to obtain the
final multiclass prediction.

\section{Experimental Setup}
\label{sec:experimental_setup}

The experimental setup was designed to evaluate the proposed framework
under severe class imbalance and across arrival-airport contexts,
with a comparison of pooled and airport-specific learning strategies.
This section first describes the flight dataset, landing-phase
windows, event labels, and retained input variables. It then presents
the preprocessing and feature-construction procedure, benchmark and
training protocols, model-selection strategy, and evaluation criteria.

\subsection{Dataset and Safety-Event Definition}
\label{subsec:dataset_definition}

\subsubsection{Dataset Source}
\label{subsubsec:dataset_source}

The experiments used the \emph{Curated 4 Class Anomaly Detection Data
Set} distributed through the NASA DASHlink Flight Data
project~\cite{dashlink_fourclass}. The underlying records consist of
de-identified flight data collected onboard a single type of regional
jet operating in commercial service over a three-year period. The
publicly released records contain aircraft-state, system-performance,
and operational variables, but do not identify the airline or aircraft
manufacturer. NASA specifies that these records are not part of an
airline Flight Operational Quality Assurance program; they are provided
as a public resource for developing and evaluating aviation-safety
data-mining methods~\cite{dashlink_sample_flight_data}.

The curated dataset contains approximately $99{,}000$ final-approach
windows. Each observation consists of a $160$-s multivariate sequence
aligned with the aircraft crossing approximately $1{,}000$~ft before
touchdown and is assigned to one nominal class or one of three
safety-event classes~\cite{dashlink_fourclass}. In the processed dataset
used in this study, the flight windows are represented as
\begin{equation}
    \mathbf{X}
    =
    \left(
        \mathbf{X}_i
    \right)_{i=1}^{N},
    \qquad
    \mathbf{X}_i
    \in
    \mathbb{R}^{T\times p},
    \label{eq:dataset_representation}
\end{equation}
where $N=99{,}837$ is the total number of available flight windows,
$T=160$ is the number of temporal samples per window, and $p=20$ is
the number of retained flight parameters. The complete input array
therefore has dimensions
\begin{equation}
    \mathbf{X}
    \in
    \mathbb{R}^{99{,}837\times160\times20}.
    \label{eq:dataset_dimensions}
\end{equation}

Each flight window is associated with a metadata record containing
the corresponding safety-event label and operational information,
including the departure and arrival airports. Because the analyzed
sequence represents the final approach to landing, the present study
defines the operational context using the \emph{arrival airport}.
The approach trajectory, runway environment, local procedures, and
surrounding operational conditions relate to the destination at which
the landing sequence occurs.

The $160$-s window provides a common temporal reference for all flights
and allows both branches of the proposed framework to operate on the
same multivariate sequence. The LSTM autoencoder uses the ordered
histories directly to characterize temporal deviations from Normal
operation, while the statistical branch summarizes each parameter
history over the same window.

\subsubsection{Airport Selection, Safety-Event Labels, and Class Distribution}
\label{subsubsec:airport_class_distribution}

The present analysis was restricted to flights arriving at
Minneapolis--Saint Paul International Airport (KMSP), Detroit
Metropolitan Wayne County Airport (KDTW), and Memphis International
Airport (KMEM). These airports were selected because they contained
comparatively large numbers of landing-phase windows while retaining
sufficient observations from the minority safety-event classes for
airport-specific model development and evaluation.

For airport
$a\in\{\mathrm{KMSP},\mathrm{KDTW},\mathrm{KMEM}\}$,
the corresponding subset is defined as
\begin{equation}
    \mathcal{D}_{a}
    =
    \left\{
        \left(\mathbf{X}_i,y_i\right)
        \;\middle|\;
        \mathrm{arrival}_i=a
    \right\},
    \label{eq:airport_specific_dataset}
\end{equation}
where $\mathbf{X}_i\in\mathbb{R}^{T\times p}$ denotes the
multivariate landing-phase window and $y_i$ denotes its associated
class label.

The dataset contains four mutually exclusive labels:
Normal, Speed High, Path High, and Flaps Late. The label set is written
as
\begin{equation}
    \mathcal{Y}
    =
    \{
        \mathrm{Normal},
        \mathrm{Speed\ High},
        \mathrm{Path\ High},
        \mathrm{Flaps\ Late}
    \}.
    \label{eq:event_label_set}
\end{equation}
Normal denotes landing windows without any of the three designated
events. Speed High corresponds to an elevated-speed condition during
approach, Path High represents an approach above the intended vertical
path, and Flaps Late indicates delayed establishment of the landing-flap
configuration. These descriptions give the operational interpretation
of the supplied labels; no event-labeling thresholds were introduced
in the present study.

Table~\ref{tab:airport_dataset_composition} summarizes the selected
arrival-airport subsets. KMSP contains the largest number of windows,
followed by KDTW and KMEM. Across all three airports, the Normal class
dominates the data, while Path High and Flaps Late contain substantially
fewer observations. Minority-class support also varies across airports.
For example, KMSP contains only 99 Flaps Late windows, whereas KMEM
contains 154 despite having a much smaller total dataset. Conversely,
the Path High class contains 272 windows at KMSP but only 61 at KMEM.

\begin{table}[t]
\centering
\caption{Arrival-airport-only dataset composition used in the pooled
and airport-specific experiments.}
\label{tab:airport_dataset_composition}
\begin{tabular}{lrrrrr}
\toprule
\textbf{Airport} &
\textbf{Normal} &
\textbf{Speed High} &
\textbf{Path High} &
\textbf{Flaps Late} &
\textbf{Total} \\
\midrule
KMSP   & 23,758 & 1,267 & 272 &  99 & 25,396 \\
KDTW   & 17,037 & 1,454 & 157 & 149 & 18,797 \\
KMEM   &  8,374 &   968 &  61 & 154 &  9,557 \\
\midrule
Pooled & 49,169 & 3,689 & 490 & 402 & 53,750 \\
\bottomrule
\end{tabular}
\end{table}

\begin{figure}[t]
    \centering
    \includegraphics[width=\columnwidth]{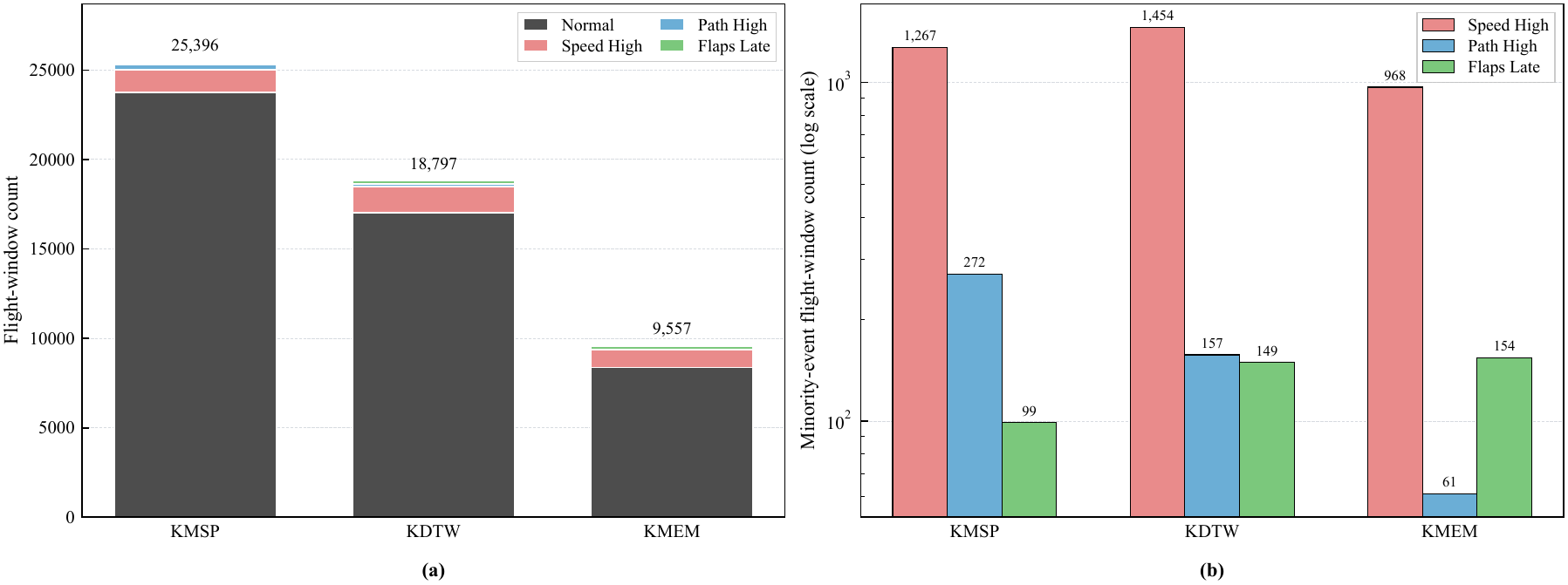}
    \caption{Arrival-airport-only dataset composition for KMSP, KDTW,
    and KMEM: (a) total class composition at each airport and
    (b) minority-event support shown using a logarithmic vertical
    scale.}
    \label{fig:airport_class_composition}
\end{figure}

Figure~\ref{fig:airport_class_composition}(a) shows the overall class
composition at each airport, while
Fig.~\ref{fig:airport_class_composition}(b) isolates the three
minority-event classes using a logarithmic vertical scale. This
separate view makes differences in event support visible despite
the predominance of the Normal class.

The pooled dataset contains 53,750 windows across the three airports,
of which 49,169 belong to the Normal class. Thus, approximately
91.5\% of the selected observations are Normal, while each individual
event class accounts for only a small fraction of the pooled data.
This imbalance motivates class-weighted XGBoost training and
class-balanced evaluation measures. The unequal event support across
airports also motivates comparison between a pooled model, which
benefits from a larger combined training set, and airport-specific
models, which may better capture local approach patterns.

\subsubsection{Input Flight Parameters}
\label{subsubsec:input_parameters}

Each landing-phase window contains the 20 recorded parameters listed in
Table~\ref{tab:input_flight_parameters}. These variables collectively
describe aircraft motion, guidance-path tracking, control-surface
response, propulsion state, landing configuration, and atmospheric
conditions, providing a multivariate description of the approach.

\begin{table}[t]
\centering
\caption{Flight parameters retained in each landing-phase window.}
\label{tab:input_flight_parameters}
\begin{tabular}{cllcll}
\toprule
\textbf{No.} & \textbf{Recorded parameter} & \textbf{Unit/type} &
\textbf{No.} & \textbf{Recorded parameter} & \textbf{Unit/type} \\
\midrule
1  & Left aileron position       & deg &
11 & Selected heading            & deg \\

2  & Right aileron position      & deg &
12 & Localizer deviation         & \% \\

3  & Corrected angle of attack   & deg &
13 & Average core speed          & \% \\

4  & Baro-corrected altitude     & ft &
14 & Total pressure              & mbar \\

5  & Computed airspeed           & kt &
15 & Pitch angle                 & deg \\

6  & Selected course             & deg &
16 & Roll angle                  & deg \\

7  & Drift angle                 & deg &
17 & Rudder position             & deg \\

8  & Left elevator position      & deg &
18 & True heading                & deg \\

9  & Trailing-edge flap position & deg/discrete &
19 & Vertical acceleration       & $g$ \\

10 & Glideslope deviation        & \% &
20 & Wind speed                  & kt \\
\bottomrule
\end{tabular}
\end{table}

The selected variables contain both event-specific and contextual
information. Computed airspeed is directly relevant to the Speed High
class, while glideslope deviation characterizes vertical-path tracking.
Trailing-edge flap position captures the timing and sequence of
landing-configuration deployment. These histories are considered
alongside altitude, angle of attack, pitch attitude, control-surface
positions, propulsion state, and atmospheric conditions.

\begin{figure*}[t]
    \centering
    \includegraphics[width=\textwidth]{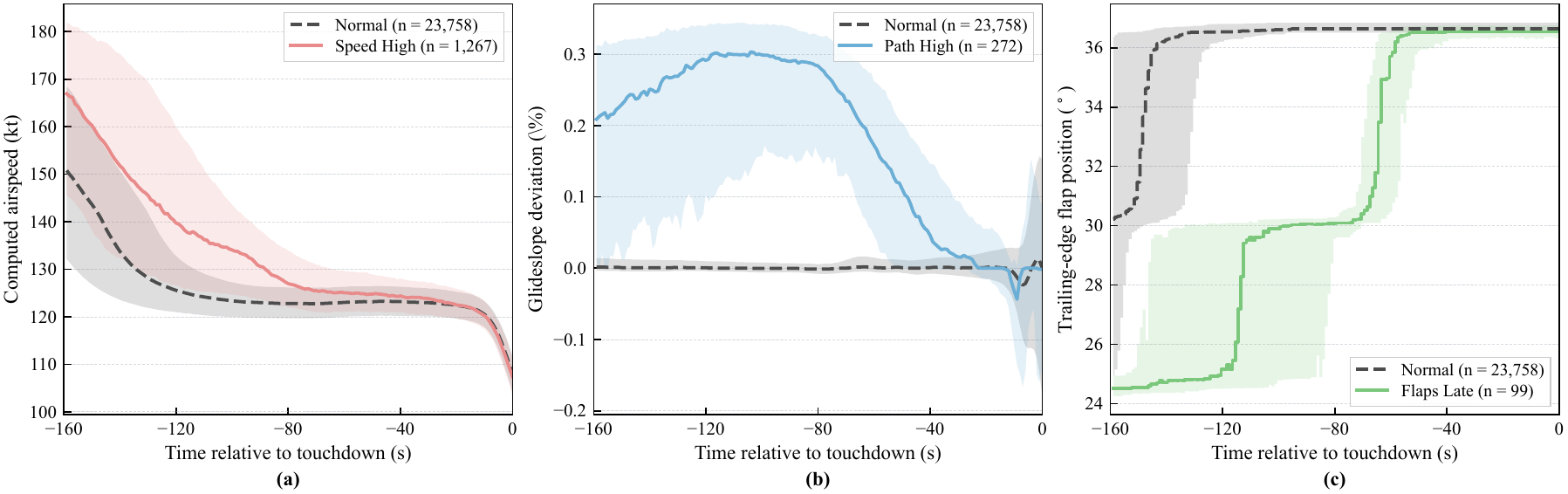}
    \caption{Class-specific landing-phase signatures at KMSP:
    (a) computed airspeed for Normal and Speed High windows,
    (b) glideslope deviation for Normal and Path High windows, and
    (c) trailing-edge flap position for Normal and Flaps Late windows.
    Curves show the class-wise median and shaded regions indicate the
    interquartile range. Time is measured relative to touchdown.}
    \label{fig:kmsp_class_signatures}
\end{figure*}

Figure~\ref{fig:kmsp_class_signatures} illustrates how the three
safety-event classes differ from Normal operation in their most
directly associated parameters at KMSP. Each panel compares an event
class with the Normal reference using the class-wise median and
interquartile range, showing both the typical approach history and
the variability within each class.

In Fig.~\ref{fig:kmsp_class_signatures}(a), Speed High windows show
greater computed airspeed through much of the approach, with the
largest median difference occurring earlier in the window. The
profiles converge close to touchdown, indicating that the distinguishing
speed-management behavior occurs over the preceding approach history.
Figure~\ref{fig:kmsp_class_signatures}(b) shows a pronounced positive
glideslope deviation for Path High windows during the earlier portion
of the approach, whereas the Normal median remains close to the
reference path. The Path High median subsequently approaches the
Normal profile as the vertical-path deviation diminishes. In
Fig.~\ref{fig:kmsp_class_signatures}(c), Normal flights typically
establish the final flap setting relatively early, while Flaps Late
windows retain intermediate settings for substantially longer.
Here, the distinguishing feature is the timing of configuration
deployment, even though both median profiles eventually reach the
same final setting.

\begin{figure*}[t]
    \centering
    \includegraphics[width=\textwidth]{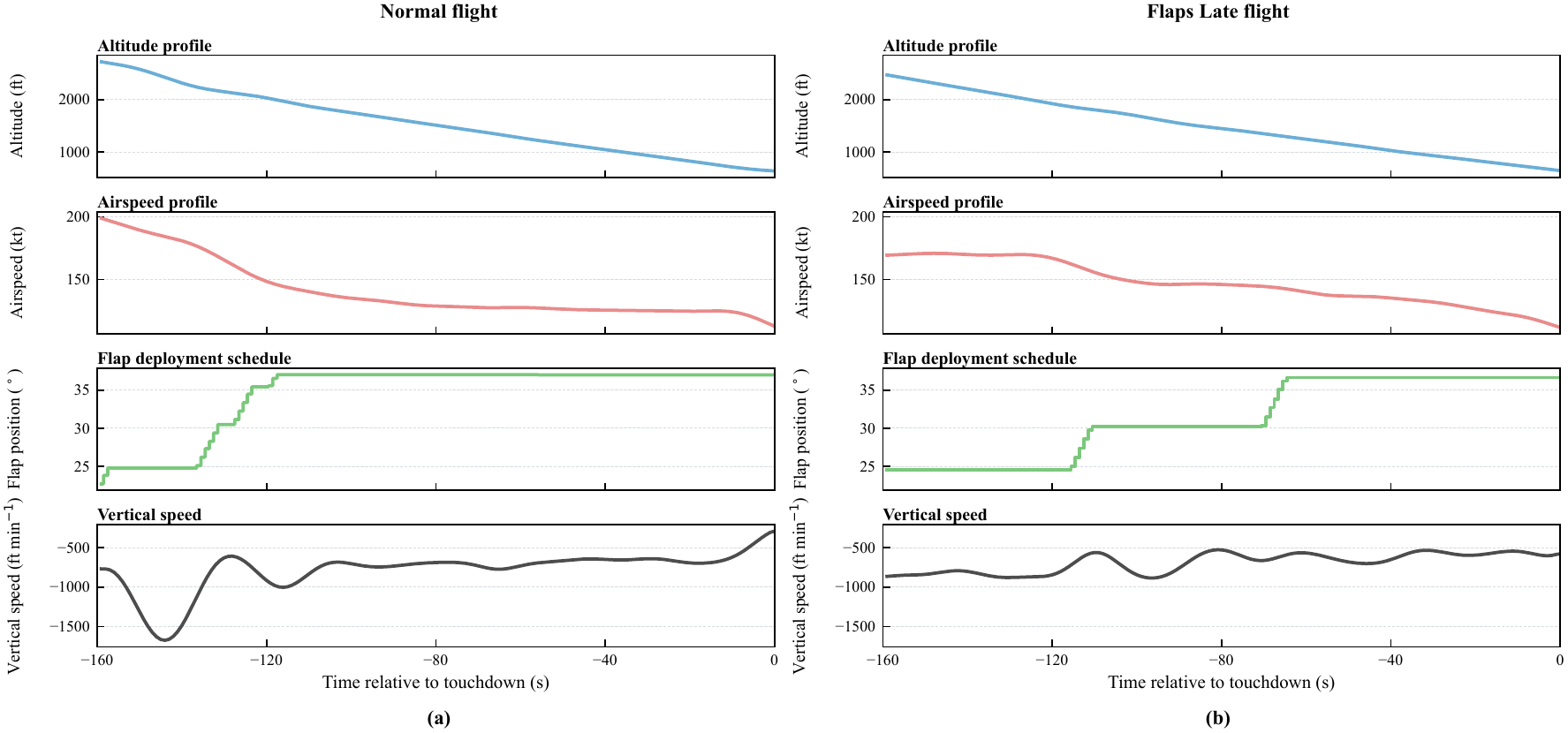}
    \caption{Illustrative multivariate landing-phase histories for
    sampled KDTW windows: (a) Normal flight and
    (b) Flaps Late flight. The four rows show altitude, computed
    airspeed, trailing-edge flap position, and vertical speed over the
    160-s interval preceding touchdown. Vertical speed is obtained by
    numerically differentiating the altitude history with respect to
    time and is shown for illustration only; it is not a model input.}
    \label{fig:representative_flights}
\end{figure*}

These aggregate profiles show the characteristic differences associated
with each event class. To examine how configuration timing appears
alongside other flight parameters within an individual approach,
Fig.~\ref{fig:representative_flights} compares one sampled Normal
flight with one Flaps Late flight from KDTW. The altitude, airspeed,
flap-position, and vertical-speed histories provide a common temporal
view of each approach.

Both examples show broadly comparable descending altitude profiles,
but their configuration and dynamic histories differ. In the Normal
example, flap deployment progresses to the final landing setting
during the earlier portion of the window, alongside a substantial
reduction in airspeed. The Flaps Late example remains at intermediate
flap settings until much later in the approach, even as the aircraft
continues to descend and reduce speed. The vertical-speed histories
also reveal differences in the descent-rate variation that are less
apparent from the altitude profiles alone. These examples illustrate
why configuration timing is considered within the evolving
multivariate approach history: similar overall descent profiles can
contain different sequences of speed reduction, flap deployment,
and descent-rate adjustment.

\subsection{Data Preprocessing and Model Configuration}
\label{subsec:model_implementation}

Following arrival-airport filtering, each flight window was represented
by a fixed-length multivariate sequence
$\mathbf{X}_i\in\mathbb{R}^{160\times20}$. The recorded parameter
histories were preprocessed to reduce high-frequency fluctuations and
account for differences in measurement scales. A third-order
Butterworth low-pass filter with a normalized cutoff frequency of
$0.1$ was applied using forward--backward filtering to suppress
high-frequency variations without introducing temporal phase shifts.
The filtered sequences were then standardized for each flight
parameter using the mean and standard deviation estimated from the
corresponding training partition. The same training-derived
standardization was applied to validation and test observations.

Feature construction followed the dual-branch formulation described
in Section~\ref{sec:methodology}. Eleven statistical descriptors
were extracted from each of the 20 parameter histories, producing
a 220-dimensional representation. In parallel, the LSTM autoencoder
was trained exclusively on Normal-class training windows and
subsequently applied to observations from all four classes. Its
mean-squared reconstruction error was retained as a continuous
scalar descriptor and concatenated with the statistical features
to form the final 221-dimensional hybrid representation.

The autoencoder used a stacked LSTM encoder--decoder architecture
with 64 units in the initial encoder layer and a 32-dimensional
latent representation. The latent state was repeated across the
temporal dimension and decoded to reconstruct the complete input
sequence. The model used a dropout rate of $0.2$, $L_2$
regularization of $10^{-3}$, and Adam optimization with an initial
learning rate of $10^{-3}$. Training was performed with a batch
size of 32 for a maximum of 210 epochs, using early stopping with
a patience of 15 epochs based on reconstruction loss on the
Normal-class validation subset.

The supervised classification stage used XGBoost with the
\texttt{multi:softprob} objective and class-dependent sample
weights defined by Eqs.~\ref{eq:class_weight}
and~\ref{eq:sample_weight}. The classifier configuration comprised
1,000 boosting iterations, a maximum tree depth of six, a learning
rate of $0.1$, and row and column subsampling ratios of $0.8$.
The model hyperparameters are summarized in
Table~\ref{tab:model_hyperparameters}. The classifier was trained
using the fused training features, with validation data used to
monitor model performance. The autoencoder parameters remained
fixed during classifier training and subsequent inference.

\begin{table}[t]
\centering
\caption{Hyperparameter configurations of the LSTM autoencoder
and class-weighted XGBoost classifier used in the proposed framework.}
\label{tab:model_hyperparameters}
\small
\setlength{\tabcolsep}{4pt}
\begin{tabular}{lclc}
\toprule
\multicolumn{2}{c}{\textbf{LSTM Autoencoder}} &
\multicolumn{2}{c}{\textbf{XGBoost Classifier}} \\
\cmidrule(lr){1-2} \cmidrule(lr){3-4}
\textbf{Hyperparameter} & \textbf{Value} &
\textbf{Hyperparameter} & \textbf{Value} \\
\midrule
Encoder LSTM units      & 64    & Objective            & \texttt{multi:softprob} \\
Latent dimension        & 32    & Boosting iterations  & 1,000 \\
Dropout rate            & 0.2   & Maximum depth        & 6 \\
$L_2$ regularization    & 0.001 & Learning rate        & 0.1 \\
Optimizer               & Adam  & Row subsampling      & 0.8 \\
Learning rate           & 0.001 & Column subsampling   & 0.8 \\
Maximum epochs          & 210   & Class weighting      & Balanced \\
Batch size              & 32    & Sample weights       & $\omega_i=w_{y_i}$ \\
Early-stopping patience & 15    & Number of classes    & 4 \\
Reconstruction loss     & MSE   &                      & \\
\bottomrule
\end{tabular}
\end{table}

\subsection{Experimental Design and Evaluation Metrics}
\label{subsec:experimental_protocol}

Two complementary training strategies were investigated to evaluate
the influence of arrival-airport context on safety-event prediction.
In the pooled strategy, a single global model was trained using
the combined KMSP, KDTW, and KMEM observations. In the
airport-specific strategy, independent specialist models were
trained using data from each airport separately. Both strategies
were evaluated on identical airport-specific test observations,
enabling direct comparison under consistent evaluation conditions.

The available observations were divided into approximately
70\% training, 10\% validation, and 20\% testing partitions.
Table~\ref{tab:arrival_airport_split} summarizes the class-wise
distributions for each airport and the pooled dataset. The pooled
partitions were constructed by combining the corresponding
airport-specific partitions. Training data were used for model
fitting, validation data for model selection, and test data
exclusively for reporting predictive performance.

\begin{table*}[t]
\centering
\caption{Training, validation, and test partitions for the three
arrival airports and the combined pooled dataset. Class-wise counts
document the support available for each safety-event class.}
\label{tab:arrival_airport_split}
\small
\begin{tabular}{llrrrrr}
\toprule
\textbf{Dataset} & \textbf{Split} &
\textbf{Total} & \textbf{Normal} &
\textbf{Speed High} & \textbf{Path High} &
\textbf{Flaps Late} \\
\midrule
\multirow{4}{*}{KMSP}
& Total      & 25396 & 23758 & 1267 & 272 & 99 \\
& Train      & 17776 & 16630 & 886  & 191 & 69 \\
& Validation & 2540  & 2376  & 127  & 27  & 10 \\
& Test       & 5080  & 4752  & 254  & 54  & 20 \\
\midrule
\multirow{4}{*}{KDTW}
& Total      & 18797 & 17037 & 1454 & 157 & 149 \\
& Train      & 13157 & 11925 & 1018 & 110 & 104 \\
& Validation & 1880  & 1704  & 145  & 16  & 15 \\
& Test       & 3760  & 3408  & 291  & 31  & 30 \\
\midrule
\multirow{4}{*}{KMEM}
& Total      & 9557 & 8374 & 968 & 61 & 154 \\
& Train      & 6689 & 5861 & 677 & 43 & 108 \\
& Validation & 956  & 838  & 97  & 6  & 15 \\
& Test       & 1912 & 1675 & 194 & 12 & 31 \\
\midrule
\multirow{4}{*}{Pooled}
& Total      & 53750 & 49169 & 3689 & 490 & 402 \\
& Train      & 37622 & 34416 & 2581 & 344 & 281 \\
& Validation & 5376  & 4918  & 369  & 49  & 40 \\
& Test       & 10752 & 9835  & 739  & 97  & 81 \\
\bottomrule
\end{tabular}
\end{table*}

To assess the contributions of the major framework components,
five benchmark configurations were considered, as summarized in
Table~\ref{tab:benchmarking_models}. These comparisons examine
the effects of class weighting, the discriminative value of the
autoencoder reconstruction error, and the distinction between
continuous feature fusion and hard-threshold anomaly gating.
The component-level experiments were performed under the
airport-specific specialist setting using the same training,
validation, and test partitions. For M5, the anomaly threshold was set to the 95th percentile
of Normal-class training reconstruction errors. Windows at or
below this threshold were assigned to Normal; threshold-exceeding
windows were classified using an event-only, class-weighted
XGBoost model trained on the 220 statistical features.

\begin{table*}[t]
\centering
\caption{Benchmark and ablation configurations used to assess
the contributions of class weighting, autoencoder-based temporal
deviation, and hybrid feature fusion.}
\label{tab:benchmarking_models}
\small
\begin{tabularx}{\textwidth}{c>{\raggedright\arraybackslash}X>{\raggedright\arraybackslash}X}
\toprule
\textbf{Model} &
\textbf{Feature input / decision logic} &
\textbf{Purpose} \\
\midrule
M1 & Statistical features + unweighted XGBoost
   & Baseline classifier \\
M2 & Statistical features + weighted XGBoost
   & Effect of class weighting \\
M3 & AE reconstruction error + weighted XGBoost
   & AE-error-only baseline \\
\textbf{M4} & \textbf{Statistical features + AE reconstruction error + weighted XGBoost}
   & \textbf{Proposed hybrid fusion} \\
M5 & Hard AE gate + statistical-feature event classifier
   & Hard-threshold baseline \\
\bottomrule
\end{tabularx}
\end{table*}

To assess robustness to data partitioning and retraining, the
airport-specific hybrid models were additionally evaluated using
five random seeds.
For each run, the training, validation, and test partitions were
regenerated and the complete learning pipeline was retrained,
including the LSTM autoencoder, reconstruction-error generation,
and class-weighted XGBoost classifier. Performance was summarized
using the mean and standard deviation across the five runs.

Classification performance was evaluated using overall accuracy,
class-wise precision and recall, macro recall, and macro F1.
For class $c$, precision and recall are defined as
\begin{equation}
    P_c =
    \frac{\mathrm{TP}_c}{\mathrm{TP}_c+\mathrm{FP}_c},
    \qquad
    R_c =
    \frac{\mathrm{TP}_c}{\mathrm{TP}_c+\mathrm{FN}_c},
    \label{eq:class_precision_recall}
\end{equation}
where $\mathrm{TP}_c$, $\mathrm{FP}_c$, and $\mathrm{FN}_c$
denote the true-positive, false-positive, and false-negative
counts for class $c$, respectively. Macro recall and macro F1
assign equal weight to all four classes:
\begin{equation}
    \mathrm{MR}
    =
    \frac{1}{C}\sum_{c=1}^{C}R_c,
    \qquad
    \mathrm{MF1}
    =
    \frac{1}{C}
    \sum_{c=1}^{C}
    \frac{2P_cR_c}{P_c+R_c}.
    \label{eq:macro_metrics}
\end{equation}
The complement $1-\mathrm{MR}$ was also reported as the
macro-averaged false-negative rate across all four classes,
including Normal. Row-normalized confusion matrices were used
to examine class-specific error patterns. Macro recall was
adopted as the principal evaluation metric because it assigns
equal importance to all classes regardless of their sample
frequencies.

Model interpretation was performed using Tree
SHAP~\cite{lundberg2017unified,lundberg2020local}.
SHAP values were computed for the 221-dimensional fused
representation and examined separately for each class output.
The absolute SHAP values associated with the 11 statistical
descriptors of each flight parameter were aggregated to obtain
parameter-level importance scores, while the autoencoder
reconstruction-error descriptor was retained separately.
For visualization, these scores were normalized by the maximum
importance within each airport--class combination. The normalized
scores show the relative ranking of feature drivers within each
combination without implying direct comparability of absolute
SHAP magnitudes across models or classes.

\section{Results and Discussion}

This section evaluates the proposed framework through comparisons of
pooled and airport-specific performance, class-wise error analysis,
and component-level benchmarking. Repeated training runs assess
variability across data partitions and retraining, while SHAP-based
interpretation examines the flight parameters associated with
predictions across airports and event classes.

\subsection{Pooled versus Airport-Specific Model Performance}
\label{subsec:global_vs_local}

A central question in landing-phase safety-event prediction is
whether pooled learning across airports offers advantages over
airport-specific models. Pooled models benefit from larger and
more diverse training datasets, while specialist models may better
capture local operational characteristics. The two training
strategies and their corresponding data partitions are described
in Section~\ref{subsec:experimental_protocol}.

Table~\ref{tab:global_vs_specialist} compares the pooled global
model with the airport-specific specialist models on identical
arrival-airport test sets. In addition to accuracy, macro recall
is reported to give equal weight to all four classes despite
their unequal sample frequencies. The percentage shown in
parentheses beside each global macro-recall value represents its
relative change with respect to the corresponding specialist model:
\begin{equation}
\Delta_{\mathrm{MR}} =
\frac{\mathrm{MR}_{\mathrm{global}}-
\mathrm{MR}_{\mathrm{specialist}}}
{\mathrm{MR}_{\mathrm{specialist}}}
\times 100 .
\label{eq:relative_macro_recall}
\end{equation}
A positive value indicates an advantage for the pooled model,
whereas a negative value indicates an advantage for the
airport-specific model.

\begin{table}[t]
\centering
\caption{Comparison of pooled global and airport-specific
specialist models under arrival-airport-only evaluation.
Parenthetical percentages indicate the relative change in
global macro recall compared with the specialist model.}
\label{tab:global_vs_specialist}
\small
\setlength{\tabcolsep}{4pt}
\begin{tabular}{lcccccc}
\toprule
\textbf{Airport} &
\textbf{Test} &
$\mathbf{Acc.}_{\mathrm{spec}}$ &
$\mathbf{Acc.}_{\mathrm{global}}$ &
$\mathbf{MR}_{\mathrm{spec}}$ &
$\mathbf{MR}_{\mathrm{global}}$ &
$\boldsymbol{1-\mathrm{MR}}_{\mathrm{spec/global}}$ \\
\midrule
KMSP & 5080 & 0.9856 & 0.9852 &
0.9572 & 0.9474 {\scriptsize(-1.02\%)} &
0.0428 / 0.0526 \\

KDTW & 3760 & 0.9779 & 0.9771 &
0.8950 & 0.8886 {\scriptsize(-0.72\%)} &
0.1050 / 0.1114 \\

KMEM & 1912 & 0.9702 & 0.9754 &
0.8599 & 0.8818 {\scriptsize(+2.55\%)} &
0.1401 / 0.1182 \\
\bottomrule
\end{tabular}
\end{table}

Both modeling strategies achieve high overall accuracy across
the three arrival airports, but macro recall reveals differences
in their performance across classes. The specialist models achieve
higher macro recall at KMSP and KDTW, where the global model is
lower by 1.02\% and 0.72\%, respectively, relative to the
corresponding specialist. At KMEM, the global model performs
better, with macro recall 2.55\% higher than that of the specialist.

The airport-dependent advantage is consistent with the tradeoff
between local specialization and training-data availability.
KMSP and KDTW have larger local datasets, which may provide
stronger support for learning airport-specific approach patterns.
At KMEM, the local subset is smaller and the Path High class
is particularly underrepresented, with only 43 training samples
and 12 test samples (Table~\ref{tab:arrival_airport_split}).
The pooled model's higher macro recall at this airport suggests
that access to additional training observations can be more
beneficial than local specialization when event support is limited.
These results support evaluating the choice of training strategy
for each airport rather than assuming a uniform benefit from
airport-specific modeling.

The following subsection examines class-wise recall and confusion
patterns for both strategies to identify which event classes
contribute to these airport-level differences.

\subsection{Class-Wise Error Behavior Across Airports}
\label{subsec:classwise_error_behavior}

\begin{figure*}[t]
    \centering
    \includegraphics[width=\textwidth]{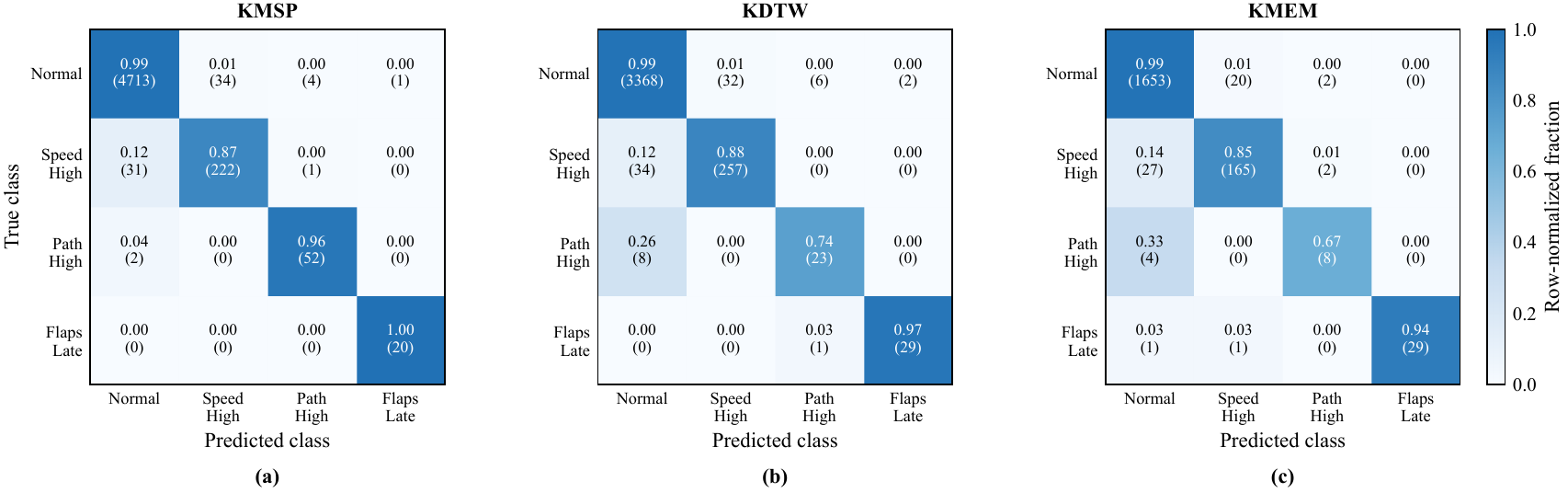}
    \caption{Row-normalized confusion matrices for the airport-specific
    specialist models under arrival-airport-only evaluation. Each cell
    reports the fraction of samples in the true class assigned to the
    predicted class, with the corresponding raw count shown in
    parentheses.}
    \label{fig:cm_specialist}
\end{figure*}

\begin{figure*}[t]
    \centering
    \includegraphics[width=\textwidth]{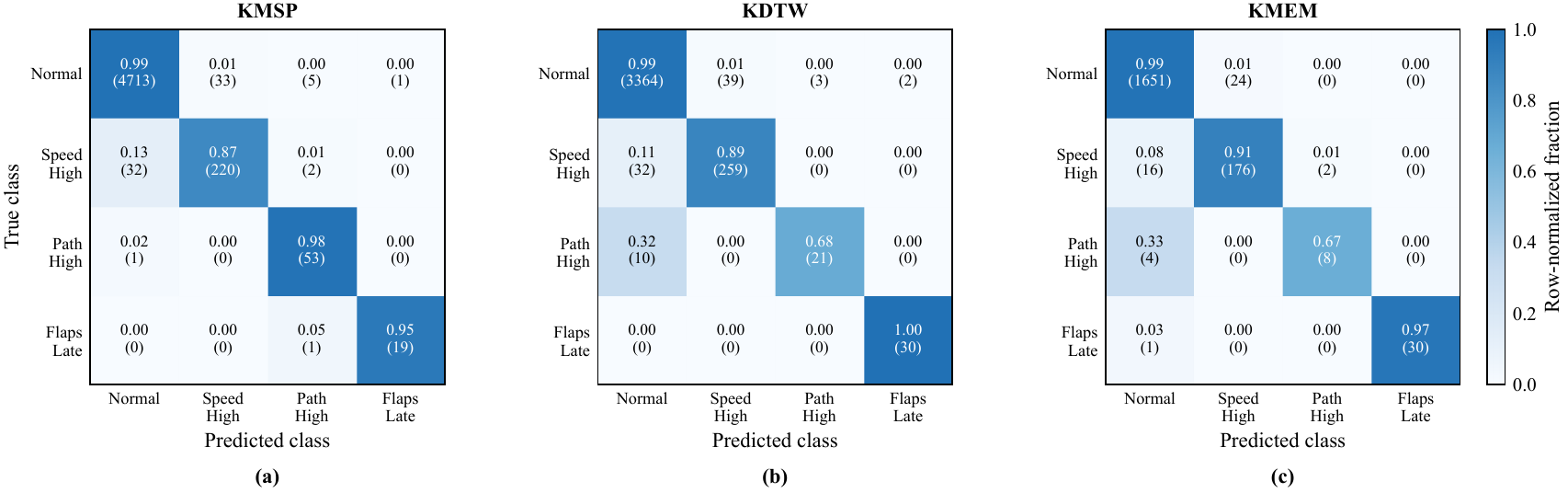}
    \caption{Row-normalized confusion matrices for the pooled global
    model evaluated on the same airport-specific arrival-only test
    sets. Each cell reports the fraction of samples in the true class
    assigned to the predicted class, with the corresponding raw count
    shown in parentheses.}
    \label{fig:cm_global}
\end{figure*}

The aggregate metrics in Table~\ref{tab:global_vs_specialist}
summarize overall performance, but do not reveal how errors are
distributed across event classes. In imbalanced flight-safety
prediction, high overall accuracy can coexist with missed detections
in rare safety-relevant classes.
Figures~\ref{fig:cm_specialist} and~\ref{fig:cm_global} therefore
show row-normalized confusion matrices for the airport-specific
specialist models and the pooled global model, respectively.
Each cell reports the fraction of samples in the true class assigned
to a predicted class, with the corresponding number of flight
windows shown in parentheses.

For the airport-specific models, the Normal class is identified
consistently across all three airports, with recall values close
to 0.99. The specialist models also show strong detection of the
Flaps Late class despite its smaller support. The KMSP specialist
correctly identifies all 20 Flaps Late cases, while the KDTW and
KMEM specialists correctly identify 29 of 30 and 29 of 31 cases,
respectively. This high recall is consistent with a distinctive
configuration-timing signature in the selected feature representation.

Path High has the lowest recall at KDTW and KMEM. The KMSP
specialist correctly identifies 52 of 54 Path High cases, giving
a recall of 0.963. In contrast, the KDTW specialist identifies
23 of 31 cases, while the KMEM specialist identifies 8 of 12.
These lower values reduce macro recall because each class
contributes equally to the macro average. All missed Path High
cases are assigned to Normal rather than to another event class,
showing that the principal difficulty for this class is separating
vertical-path deviations from nominal approaches.

Comparison with the pooled global model explains the airport-level
differences in Table~\ref{tab:global_vs_specialist}. At KMSP,
the global model produces a similar error pattern. It identifies
53 of 54 Path High cases, compared with 52 of 54 for the specialist,
but detects one fewer Flaps Late case and slightly fewer Speed
High cases. These losses outweigh the improvement in Path High
recall. At KDTW, the global model slightly improves detection of
Speed High and Flaps Late, but correctly identifies only 21 of
31 Path High cases, compared with 23 of 31 for the specialist.
The lower Path High recall accounts for the specialist model's
small macro-recall advantage at this airport.

At KMEM, the pooled model's macro-recall advantage is not due
to better Path High detection: both models identify 8 of 12 cases.
The improvement instead comes from Speed High and Flaps Late.
The global model correctly identifies 176 of 194 Speed High cases,
compared with 165 of 194 for the specialist, and 30 of 31 Flaps
Late cases, compared with 29 of 31. Thus, the benefit of pooling
at KMEM is class-dependent and does not resolve the weakest
Path High performance. 

Across both strategies, event-to-Normal confusion remains an
important source of missed detections, with particularly low
Path High recall at KDTW and KMEM. The small Path High test
sets, especially the 12 cases at KMEM, also limit the precision
of these class-wise estimates. These patterns explain why the
pooled-versus-specialist comparison requires class-sensitive
metrics and inspection of individual event classes alongside
overall accuracy.

\subsection{Robustness Analysis}
\label{subsec:random_seed_robustness}

For brevity, this and the subsequent analyses focus on the
airport-specific specialist models. Both strategies achieved
comparable overall accuracy, with modest specialist macro-recall
advantages at KMSP and KDTW, the two airports with larger local
datasets, and a pooled-model advantage at KMEM.

\begin{figure}[t]
    \centering
    \includegraphics[width=0.75\linewidth]
    {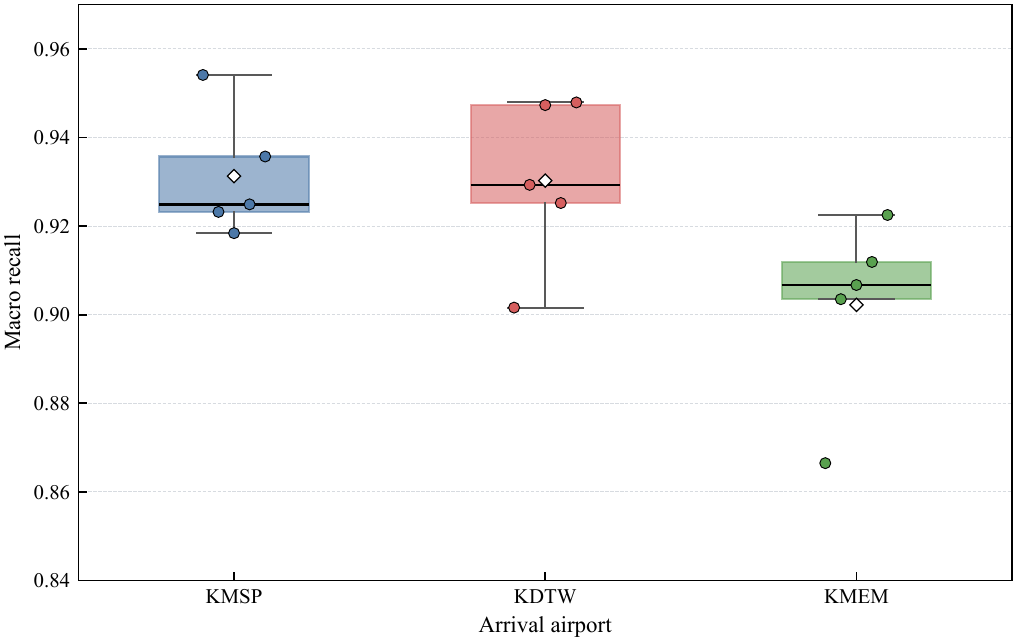}
    \caption{Macro-recall variability of the proposed airport-specific
    hybrid-fusion models across five random data splits and
    retraining runs. Individual points represent the five runs
    for each arrival airport, and diamonds denote their means.}
    \label{fig:random_seed_macro_recall}
\end{figure}

The robustness of the proposed hybrid-fusion framework was
assessed using the repeated-split and retraining protocol
described in Section~\ref{subsec:experimental_protocol}.
This analysis examines how specialist performance varies when
the training, validation, and test observations are reassigned
and the complete pipeline is retrained, particularly when
safety-event classes have limited support.
Table~\ref{tab:random_seed_robustness} reports the mean and
standard deviation of macro recall and macro F1 across the
five runs, while Fig.~\ref{fig:random_seed_macro_recall}
illustrates the corresponding variation in macro recall.

\begin{table}[t]
\centering
\caption{Performance of the proposed airport-specific hybrid-fusion
models across five random data splits and retraining runs.
Values are reported as mean $\pm$ standard deviation.}
\label{tab:random_seed_robustness}
\begin{tabular}{lcc}
\toprule
\textbf{Airport} &
\textbf{Macro Recall} &
\textbf{Macro F1} \\
\midrule
KMSP & $0.9313 \pm 0.0142$ & $0.9347 \pm 0.0062$ \\
KDTW & $0.9303 \pm 0.0190$ & $0.9229 \pm 0.0129$ \\
KMEM & $0.9022 \pm 0.0212$ & $0.9010 \pm 0.0242$ \\
\bottomrule
\end{tabular}
\end{table}

The framework maintains a mean macro recall above 0.90 for
all three airports. KMSP and KDTW exhibit comparable mean
performance, with KMSP showing the lowest standard deviation.
KMEM has a slightly lower mean macro recall and the greatest
standard deviation across the five runs. This pattern is
consistent with its smaller local dataset and limited
minority-class support, although the analysis does not isolate
the contribution of individual event classes to run-to-run
variability.

These findings support the consistency of class-balanced
performance across the evaluated partitions and retraining runs.
The greater variability at KMEM nevertheless shows that the
reported performance remains sensitive to sample allocation
and retraining when event support is limited. Airport-level
results should therefore be interpreted alongside both the
mean performance and its variation across runs.

\subsection{Component-Level Benchmarking of the Proposed Framework}
\label{subsec:benchmarking_results}

The component-level benchmarking results assess the contribution
of the major design choices in the proposed framework. The five
model configurations (M1--M5), defined in
Table~\ref{tab:benchmarking_models}, were evaluated under the
airport-specific specialist setting for KMSP, KDTW, and KMEM,
following the protocol described in
Section~\ref{subsec:experimental_protocol}.
The comparison examines the effects of class weighting, the
discriminative value of the autoencoder reconstruction error,
and the distinction between continuous feature fusion and
hard-threshold anomaly gating.

\begin{table*}[t]
\centering
\caption{Benchmarking comparison of model variants under the
arrival-airport-only specialist setting. The event-class recall
average assigns equal weight to Speed High, Path High, and
Flaps Late, excluding Normal.}
\label{tab:benchmarking_results}
\resizebox{\textwidth}{!}{%
\begin{tabular}{llccccc}
\toprule
\textbf{Airport} & \textbf{Model} & \textbf{Accuracy} &
\textbf{Macro Recall} & \textbf{Macro F1} &
\(\mathbf{1-\mathrm{MR}}\) & \textbf{Event-Class Recall Avg.} \\
\midrule
\multirow{5}{*}{KMSP}
& M1 & 0.9852 & 0.8939 & 0.9249 & 0.1061 & 0.8601 \\
& M2 & 0.9837 & 0.9529 & 0.9389 & 0.0471 & 0.9404 \\
& M3 & 0.6364 & 0.4815 & 0.2883 & 0.5185 & 0.4264 \\
& M4 & 0.9848 & 0.9542 & 0.9400 & 0.0458 & 0.9417 \\
& M5 & 0.9000 & 0.6186 & 0.6517 & 0.3814 & 0.5103 \\
\midrule
\multirow{5}{*}{KDTW}
& M1 & 0.9766 & 0.8808 & 0.9121 & 0.1192 & 0.8437 \\
& M2 & 0.9777 & 0.8934 & 0.8969 & 0.1066 & 0.8616 \\
& M3 & 0.5944 & 0.3939 & 0.2565 & 0.6061 & 0.3179 \\
& M4 & 0.9782 & 0.9046 & 0.8997 & 0.0954 & 0.8770 \\
& M5 & 0.8763 & 0.5384 & 0.5820 & 0.4616 & 0.4033 \\
\midrule
\multirow{5}{*}{KMEM}
& M1 & 0.9665 & 0.8486 & 0.8657 & 0.1514 & 0.8021 \\
& M2 & 0.9686 & 0.9008 & 0.8753 & 0.0992 & 0.8731 \\
& M3 & 0.6156 & 0.4401 & 0.2900 & 0.5599 & 0.3704 \\
& M4 & 0.9686 & 0.8662 & 0.8664 & 0.1338 & 0.8266 \\
& M5 & 0.8515 & 0.5035 & 0.5455 & 0.4965 & 0.3566 \\
\bottomrule
\end{tabular}%
}
\end{table*}

\begin{figure}[t]
    \centering
    \includegraphics[width=\linewidth]
    {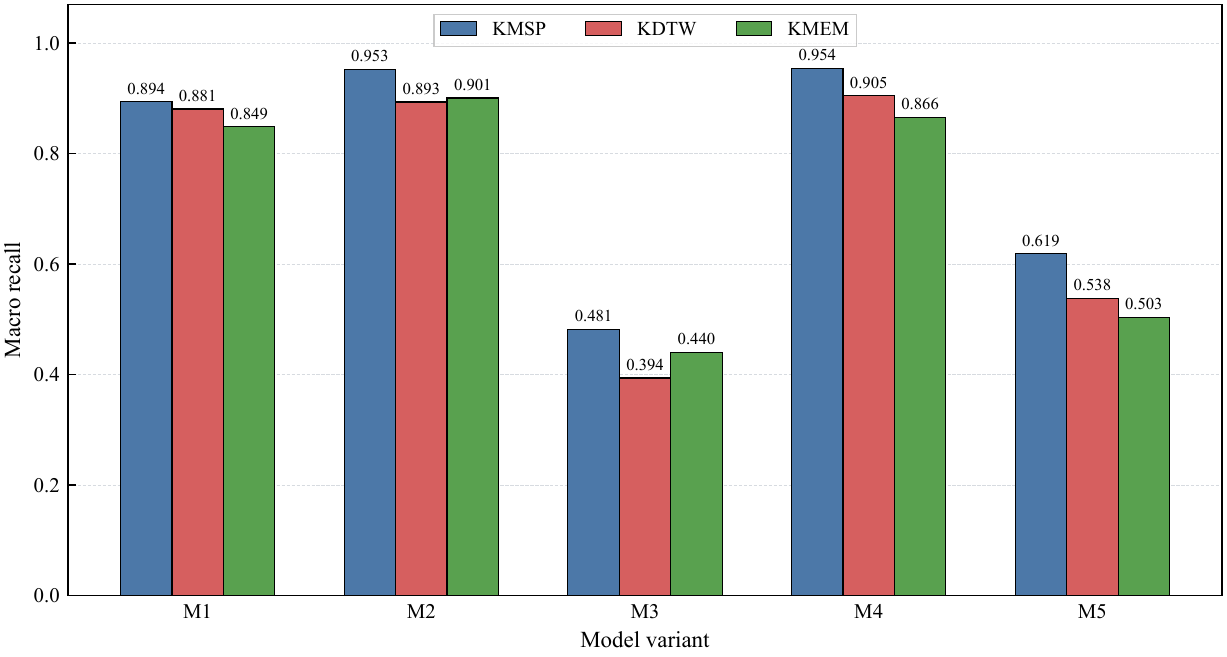}
    \caption{Macro-recall comparison of the benchmark model
    variants across the three arrival airports. Model definitions
    are given in Table~\ref{tab:benchmarking_models}.}
    \label{fig:benchmarking_macro_recall}
\end{figure}

Table~\ref{tab:benchmarking_results} and
Fig.~\ref{fig:benchmarking_macro_recall} show that the proposed
M4 model achieves the highest macro recall at KMSP and KDTW
and exceeds the unweighted, AE-error-only, and hard-gate
baselines on this metric at all three airports.
The component comparisons also establish the importance of
class weighting. At KMSP, macro recall increases from 0.8939
in M1 to 0.9529 in M2. The same trend is observed at KDTW
and KMEM, where macro recall increases from 0.8808 to
0.8934 and from 0.8486 to 0.9008, respectively. These gains
demonstrate the contribution of imbalance-aware training,
which is less apparent from the consistently high overall
accuracy of the statistical-feature models.

The AE-error-only model, M3, performs poorly across all three
airports. Its macro recall is 0.4815 for KMSP, 0.3939 for
KDTW, and 0.4401 for KMEM. A scalar reconstruction error
summarizes the degree of deviation from learned nominal
temporal behavior, but provides insufficient information on
its own to reliably distinguish among Speed High, Path High,
and Flaps Late. This result supports its role as a complementary
feature interpreted jointly with parameter-specific flight-history
descriptors.

The hard-gate baseline, M5, is also substantially weaker
than the feature-fusion formulation. In M5, samples below
the AE threshold are directly assigned to Normal, while
only threshold-exceeding samples are passed to the event
classifier. Any event assigned to Normal at the gate therefore
cannot be recovered by the downstream classifier.
Macro recall falls to 0.6186 for KMSP, 0.5384 for KDTW,
and 0.5035 for KMEM, accompanied by markedly lower
event-class recall averages. The advantage of M4 over this
baseline supports retaining reconstruction error as a
continuous classifier input, allowing its relevance to be
learned jointly with the statistical descriptors.

\begin{figure*}[t]
    \centering
    \includegraphics[width=\textwidth]
    {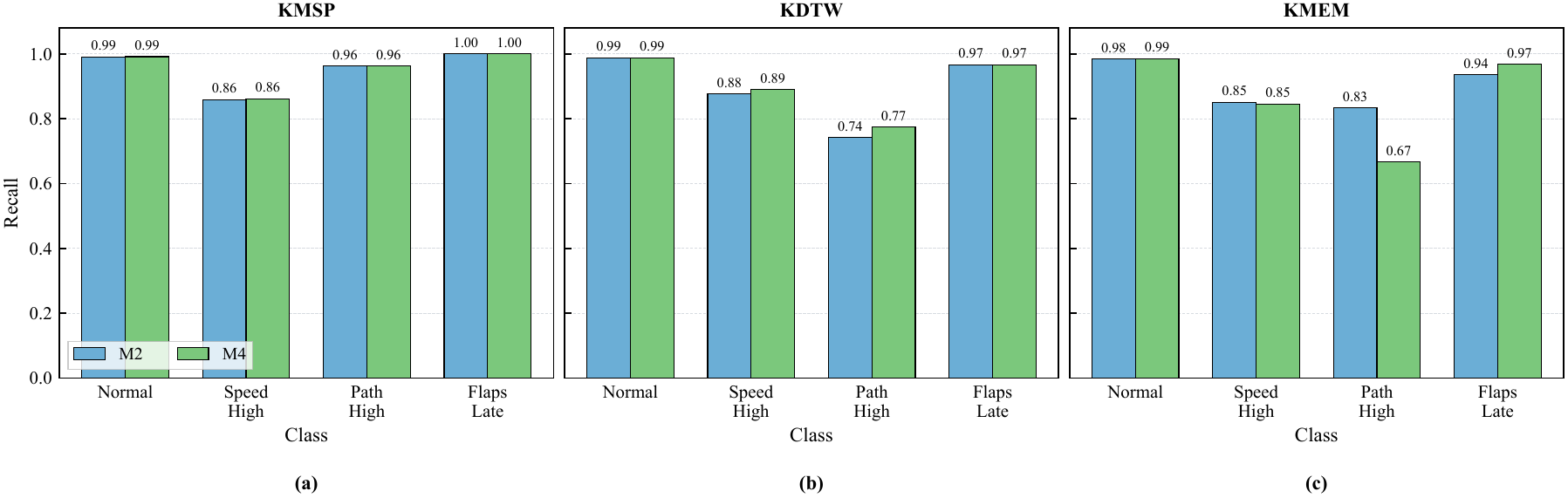}
    \caption{Class-wise recall comparison between the weighted
    statistical-feature model (M2) and the proposed hybrid-fusion
    model (M4).}
    \label{fig:benchmarking_classwise_m2_m4}
\end{figure*}

Comparison with M2 isolates the additional contribution of
the temporal-deviation descriptor within the weighted
statistical-feature classifier. M4 improves both macro recall
and macro F1 at KMSP and KDTW, while matching or improving
M2's overall accuracy at all three airports. At KMSP,
macro recall increases from 0.9529 to 0.9542; at KDTW,
it increases from 0.8934 to 0.9046. The event-class recall
average also improves at both airports, showing that the
gains extend to the safety-event classes rather than arising
solely from Normal classification. M2 retains higher macro
recall at KMEM. Figure~\ref{fig:benchmarking_classwise_m2_m4}
examines the class-wise differences underlying this exception.

The class-wise comparison shows that M4 broadly maintains
or improves the detection performance of M2, with the main
reduction concentrated in Path High at KMEM. At KMSP,
the class-wise recalls are closely matched, while at KDTW,
M4 improves Speed High and Path High detection with little
change in the remaining classes. At KMEM, M4 achieves
higher Normal and Flaps Late recall and comparable Speed
High recall, but its Path High recall decreases from 0.8333
to 0.6667. With only 12 Path High test windows, this difference
corresponds to 10 correctly classified cases for M2 and
8 for M4. The lower KMEM macro recall therefore reflects
a specific weakness in this low-support class, rather than
a broad deterioration across event types. It is also consistent
with the earlier class-wise analysis, which identified Path
High detection at KMEM as a remaining difficulty.

Taken together, the benchmarks support the proposed combination
of parameter-specific statistical descriptors, imbalance-aware
training, and continuous temporal-deviation information.
M4 achieves the best macro recall at two airports and improves
on the unweighted, AE-error-only, and hard-gate alternatives
at all three. Its favorable class-wise comparison with M2
demonstrates the complementary value of the reconstruction-error
descriptor, while the KMEM Path High result identifies a
localized limitation requiring stronger event support and
further evaluation.

\subsection{Model Explainability and Safety-Relevant Feature Drivers}
\label{subsec:shap_explainability}

To examine whether the final classifier relied on operationally
meaningful variables, Tree SHAP analysis was applied to the trained
airport-specific hybrid-fusion XGBoost models. SHAP values were
computed for the 221-dimensional fused representation, and the
statistical-descriptor attributions associated with each flight
parameter were aggregated to obtain parameter-level importance
scores. The LSTM-autoencoder reconstruction error was retained
separately. These attributions explain the contribution of the
inputs to each trained model's class-specific output.

\begin{figure*}[t]
    \centering
    \includegraphics[width=\textwidth]{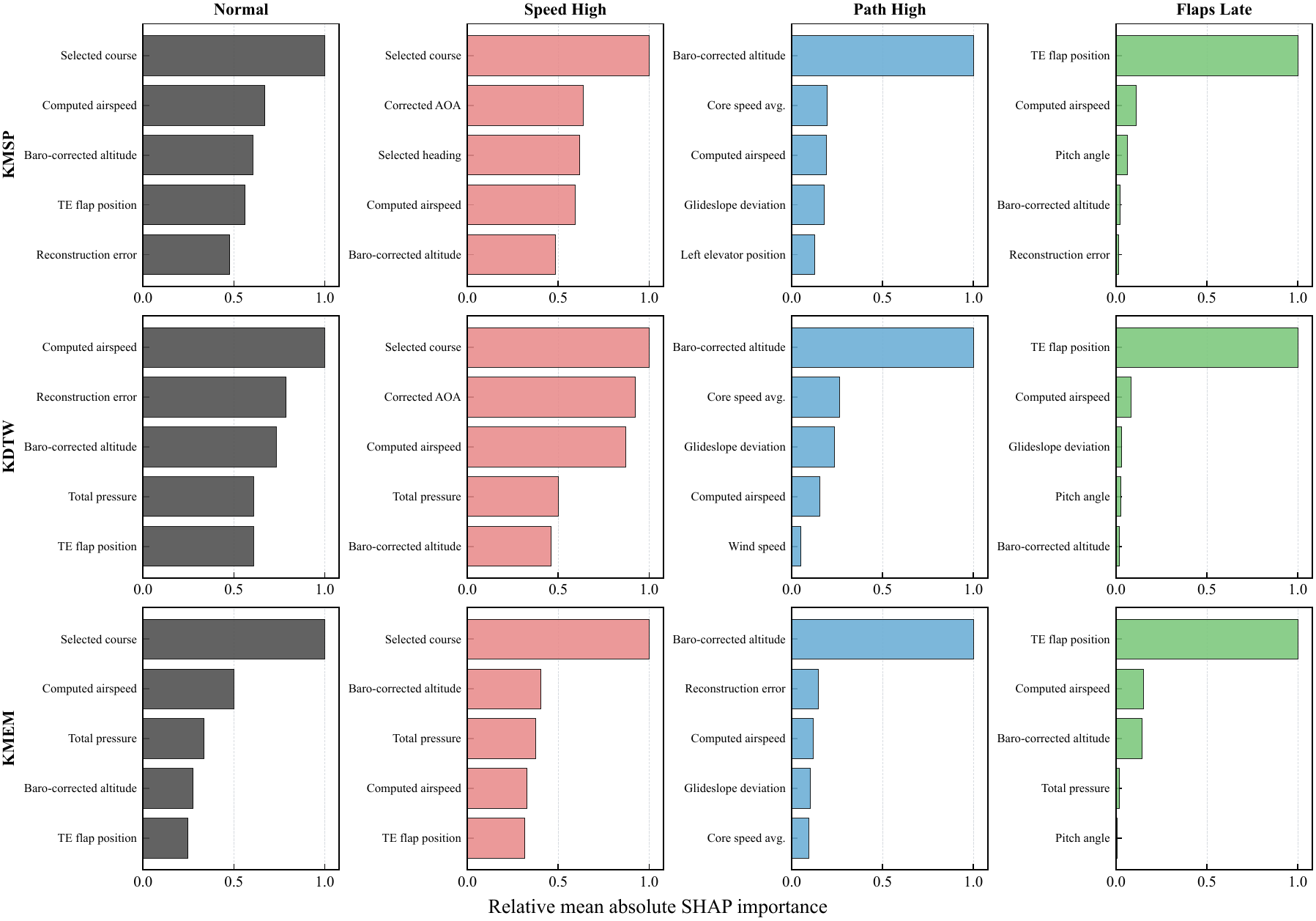}
    \caption{Top five parameter-level SHAP drivers for each class
    and arrival airport, with reconstruction error retained as a
    separate feature. Values are normalized within each airport--class
    panel by the largest aggregated mean absolute SHAP value in that
    panel, emphasizing relative importance within each panel.}
    \label{fig:shap_sensor_top5}
\end{figure*}

Figure~\ref{fig:shap_sensor_top5} shows the top five parameter-level
drivers for each class and airport, including reconstruction error
where applicable. Values are normalized within each airport--class
panel by the largest aggregated mean absolute SHAP value in that
panel. The figure therefore emphasizes relative importance within
each prediction context, without implying direct comparability of
absolute SHAP magnitudes across airports or class outputs.

The clearest alignment between feature importance and the operational
meaning of the labels occurs for Path High and Flaps Late.
For Path High, baro-corrected altitude is the dominant driver
across all three airports, while glideslope deviation consistently
appears among the top-ranked variables. This agrees with the
interpretation of Path High as a vertical-path deviation class.
Secondary drivers include computed airspeed, average core speed,
wind speed, and elevator position, showing that the class output
also uses information about speed, propulsion, atmospheric
conditions, and control response alongside the vertical-path
variables.

For Flaps Late, trailing-edge flap position dominates the
attributions at all three airports by a large margin. This
ranking aligns directly with the configuration-timing meaning
of the event label. Secondary drivers include computed airspeed,
pitch angle, altitude, glideslope deviation, and total pressure.
Their presence is operationally plausible because flap deployment
during approach accompanies changes in speed, attitude, and
flight path. The strong reliance on flap-position information
is also consistent with the high Flaps Late recall observed
in the class-wise results despite the class's relatively
small sample support.

For Speed High, selected course is the dominant driver across
all three airport-specific models. Computed airspeed, corrected
angle of attack, total pressure, and baro-corrected altitude
also appear among the important variables, linking the class
output to speed- and energy-related flight histories.
Selected course may additionally encode differences in approach
direction and local operating context. Its prominence is
consistent with the classifier using contextual associations
alongside the flight-state variables, although the specific
runway or procedural relationships underlying this association
are not established by the feature rankings.

For Normal, the leading drivers include selected course,
computed airspeed, baro-corrected altitude, flap position,
total pressure, and reconstruction error. The class output
therefore draws on several aspects of the approach history,
rather than a single event-specific parameter. Reconstruction
error ranks among the top five drivers for Normal at KMSP
and KDTW, and is also the second-ranked driver for Path High
at KMEM. These attributions show that the temporal-deviation
descriptor is used alongside the parameter-specific statistical
features. Its role is complementary, consistent with the
benchmarking results showing that reconstruction error alone
is insufficient for reliable multiclass prediction.

Across airports, the dominant drivers for vertical-path and
configuration events align with the corresponding operational
variables, while Speed High predictions also rely strongly
on approach-context information. Airport-dependent differences
in the secondary drivers show how the trained models use
different combinations of flight-history information for the
same class output. This class-specific interpretation complements
the performance comparisons by identifying the variables
associated with the models' predictions.

\section{Conclusion}
\label{sec:conclusion}

This study developed and demonstrated a hybrid feature-fusion
framework for landing-phase safety-event prediction under severe
class imbalance. The framework combines learned deviations from
nominal temporal behavior with parameter-specific flight-history
descriptors, enabling a class-weighted XGBoost classifier to distinguish
safety-event signatures across different airport contexts. Its
contribution includes the predictive framework, empirical assessment
of the value of temporal feature fusion, and evidence of how
airport-specific learning affects event detection. Class-wise
evaluation and SHAP analysis connect these performance findings
to identifiable operational variables, providing an interpretable
basis for examining the resulting predictions.

The component benchmarks demonstrate the effectiveness of the
proposed design. The hybrid model achieved the highest macro recall
at KMSP and KDTW, reaching 0.9542 and 0.9046, respectively, and
outperformed the unweighted, AE-error-only, and hard-gate baselines
on this metric at all three airports. Adding the temporal-deviation
descriptor to the weighted statistical-feature classifier improved
both macro recall and macro F1 at two airports, demonstrating its
complementary predictive value. Continuous feature fusion also
substantially outperformed hard anomaly gating, supporting the
use of reconstruction error as information for multiclass
discrimination. Mean macro recall remained above 0.90 at every
airport across five data partitions and retraining runs. The
pooled-versus-specialist comparison identified benefits from local
specialization at KMSP and KDTW and from pooled training at KMEM,
providing evidence for selecting the training strategy according
to airport context and available event data. SHAP rankings further
showed clear alignment between the dominant predictive variables
and the operational meaning of vertical-path and configuration events.

The evaluation was limited to a curated dataset from a single
regional-jet type and three arrival airports. Generalization across
aircraft types and broader operating environments therefore remains
to be established. Minority-event support was also limited,
particularly for Path High at KMEM, making its estimated recall
sensitive to individual test cases. The repeated-split analysis
characterized variability within the available dataset, but did not
evaluate transfer to unseen airports or changes in operating
conditions over time.

Future work will extend the evaluation to multiple aircraft types
and additional airports to assess the robustness of the learned
representations across fleets and operating contexts. Greater
coverage of rare events will support more reliable class-wise
comparisons and help determine when pooled training, local
specialization, or adaptation between operating contexts is most
effective. Evaluation on temporally separated data will further
assess whether the observed performance is maintained as operational
conditions change.

\section*{Funding}
This research received no external funding.

\section*{Data availability}

The flight data used in this study are publicly available from the
NASA DASHlink \emph{Curated 4 Class Anomaly Detection Data
Set}~\cite{dashlink_fourclass} at
\url{https://c3.ndc.nasa.gov/dashlink/resources/1018/}.
The code used to develop and evaluate the proposed framework is
available from the corresponding author upon reasonable request.

\section*{Declaration of generative AI use}

During the preparation of this work, the authors used Codex for language editing and to assist in generating Python
scripts for plotting. The authors carefully reviewed and checked all
manuscript content and generated scripts, made revisions as necessary,
and take full responsibility for the accuracy and integrity of the
published article.

% \bibliographystyle{unsrt}
% \bibliography{references}

\bibliographystyle{unsrtnat}
\bibliography{references}

\end{document}